*Research article*

# Prediction certification cannot replace explanation certification: a competence envelope for trustworthy AI under compound stress

**Nataliya Shakhovska**[1], **Ivan Izonin**[1], **Stergios-Aristoteles Mitoulis** [2]

[1]*Lviv Polytechnic National University, Lviv, Ukraine.*

[2] *Centre for Global Infrastructure Resilience, The Bartlett School of Sustainable Construction, University College London, UK*

Artificial-intelligence systems increasingly make consequential judgments – which patient is deteriorating, which building is safe to enter, whether an image is authentic – and are trusted on the strength of how accurately and confidently they predict. The safeguards that certify them are correspondingly prediction-based: accuracy, calibration and conformal coverage all measure how well a model performs. Whether such checks are sufficient to establish model trustworthiness has remained unclear. Here we prove that they cannot. We establish a separation theorem showing that a reliable model and a compromised one can be identical under every prediction-side certificate, including accuracy, calibration and coverage, yet differ arbitrarily in explanation fidelity and deployment behaviour. Detecting this failure requires access to the model's decision mechanism in addition to its predictions. We introduce the competence envelope as an operational framework that combines prediction and explanation certification into a single deployable criterion. Across diverse datasets and model classes, the proposed framework reveals failure modes that prediction-side certification alone does not capture. Certification against failures that are invisible in prediction behaviour therefore requires evidence about the model's decision mechanism as well as its outputs.

## Introduction

Across science, medicine and public life, consequential judgments are increasingly made or shaped by automated systems: which patient is deteriorating, which building is safe to re-enter after a disaster, whether an image or a claim is authentic. Under controlled evaluation these systems can be remarkably accurate, and that accuracy is real. But it is conditional. It rests on assumptions that are seldom stated because, in curated settings, they almost always hold: that data are plentiful and representative, that the world does not shift between calibration and use, that no one has tampered with the inputs, and that the machine runs on hardware that does not fail. Where the stakes are highest, these assumptions do not weaken one at a time; they break together. A diagnostic tool meets a population unlike any it was trained on; a damage-assessment system meets a disaster of a kind it has never seen, imaged by a sensor that is itself broken; an authenticity check meets a manipulation designed after it was built.

The deepest danger is not that such systems are sometimes wrong. Any judgment, human or machine, is sometimes wrong. The danger is that they are wrong *silently*. A system that has left the conditions under which it is competent does not announce the fact. It keeps issuing confident answers. And, in the part that has been most underweighted, the reasons it offers keep looking reasonable, because the cues a machine leans on need not be the cues a person would trust[1]. A convincing explanation then lends a wrong answer a false air of trustworthiness at the moment a human most needs to be warned to distrust it[2]. In building systems that can explain themselves, we have given silent failure an articulate voice.

This paper advances a single claim and traces its consequences. Our central result is a theorem. Monitoring what a model *predicts* is, within the class of functionals of its prediction behaviour, provably insufficient to certify whether it can be trusted. The *separation theorem* makes this exact: a reliable model and a compromised one can be made identical to every prediction-side certificate (conformal coverage, accuracy, calibration, confidence) while differing arbitrarily in the fidelity of their *explanations* and in their behaviour under shift. A matching lower bound shows that no monitor built from a model's predictions alone detects this class of failure above chance, and that detecting it requires reading the model's structure. Certifying the reasons a model gives therefore becomes a necessary condition for certifying against this class of failure, and the rest of this paper is evidence for that claim and its consequences. To organise the two certificates we introduce the *competence envelope*, the region of operating conditions over which a system's judgments *and the reasons it gives* are jointly trustworthy: a provably non-trivial region whose boundary is set, direction by direction, by whichever certificate fails first. The envelope contracts along four independently instrumented axes of deployment stress – distribution drift, data scarcity, adversarial contamination and resource degradation – with the prediction certificate failing under drift and the explanation certificate under scarcity and model compression, so that neither substitutes for the other. It can also be acted upon. A label-free monitor senses in real time when a system has left the envelope and degrades gracefully, preferring a verifiable "I do not know here" to a confident, well-explained error; computed from unlabelled inputs alone it anticipates deployment error up to six years ahead (Spearman $\rho = 0.90$ on climate discourse) and, used as an abstention gate, cuts error by 28%. The empirical sections test the theorem's implications across multiple datasets, model families, deployment conditions and operating stresses.

## The fragmentation of trustworthy AI

Trustworthy machine learning today is five mature literatures that rarely speak to one another. Research on distribution shift and domain adaptation detects and corrects changes between training and deployment[3,4], but treats explanation and abstention as out of scope. Work on small-data and imbalanced learning improves estimation under scarcity, but assumes a fixed distribution. Adversarial robustness certifies prediction stability against bounded perturbations[5], yet rarely connects to natural drift or to interpretability. Uncertainty quantification, selective prediction and conformal inference deliver calibrated abstention and distribution-free guarantees[6,7,8] (though the confidence they calibrate is itself known to degrade under shift[9]), but they guard *predictions*, not explanations. And explainable AI produces post-hoc attributions[10,11,12] whose own faithfulness is fragile and is essentially never certified as a function of operating conditions, a fragility that has prompted calls to prefer inherently interpretable models for the highest-stakes decisions[13].

Each toolkit silently assumes the others' problems away, and the result is the silent, well-explained failure above. The nearest existing concepts (out-of-distribution detection[14,15], selective prediction and conformal inference, certified robustness, applicability domains and model cards, trust scores) each fall short on a specific axis; what none characterises, let alone certifies, is the *joint* region of operating conditions over which both a prediction and its explanation may be trusted. That certifiable union is the gap.

This is not a gap between neighbouring methods but a *structural* one, and the separation theorem locates it precisely. Every framework in the trustworthy-machine-learning canon – statistical learnability and generalisation bounds, PAC-Bayes, calibration and proper scoring rules, selective prediction, split-conformal coverage, distribution-shift and domain-adaptation bounds, and certified adversarial robustness – is a *prediction-side* guarantee in the exact sense of Theorem 1: each is a functional of the model's input–output behaviour together with labels. By that theorem, no such guarantee, however tightened, can certify – from certification-time prediction-law information – the class of failures in which a model's explanation has drifted while its predictions are untouched; the entire prediction-side canon lies on one side of the separation, and structural methods that read weights, gradients or activations (mechanistic interpretability, gradient-based audits) lie on the other, with the explanation certificate. Explainable AI sits on that structural side but supplies no *certificate*: it produces attributions[10,11,12] whose own faithfulness is fragile and is essentially never bounded as a function of operating conditions[13].

What is missing from the literature is therefore not a better estimator but a *necessary condition* that no existing theory states: that certifying against this class of failure requires the explanation, not only the prediction. Stacking existing methods – conformal coverage, a robustness ball, an attribution map – cannot supply it, because each ingredient is prediction-side or uncertified, and because the stressors interact, so a guarantee proved under drift alone need not survive drift with scarcity. Related work does not merely lack the joint object; by Theorem 1 the prediction-side canon lies provably outside it. This is the primitive we introduce: explanation fidelity certified as a function of where a system operates, coupled to prediction reliability in one region with a provable boundary (Table 1).

Recent flagship studies sharpen the gap. Label-light detectors of unreliable generation[16] and the finding that larger, more instructable models become less reliable and confidently wrong[17] both operate on the prediction side; demonstrations that models decide on covert, spurious cues such as dialect[18] or a cost proxy standing in for health need[19], and the caution that explanations in medicine can mislead unless their faithfulness is itself assured[20], are exactly the failures our explanation certificate targets. What remains missing across this literature is a certifiable operational boundary – a region in which both a model's answer and the reason for that answer can be trusted; the competence envelope supplies that object.

**Table 1 |** How the competence envelope relates to neighbouring guarantees.

| Approach | What it certifies | What it misses | What the competence envelope adds |
|---|---|---|---|
| OOD / anomaly detection | that a single input is anomalous (binary verdict) | how reliable the prediction is; the explanation; a region | a graded, certified region and a fidelity-bounded explanation |
| Selective prediction / conformal inference | prediction coverage or selective risk | explanation fidelity; interacting, compound shift | a joint prediction-and-explanation certificate under compound stress |
| Certified robustness | prediction stability within a bounded perturbation ball | natural drift; scarcity; the explanation | natural compound shift plus a certified explanation |
| Applicability domain; model & robustness cards | where a model is expected to work (descriptive) | a guarantee; the explanation; boundary behaviour | a certifiable boundary with a prescribed action at it |
| Learnability & generalisation bounds (PAC, PAC-Bayes) | expected error of a predictor from finite samples | explanation fidelity; operating-condition dependence | a certified region for prediction and explanation jointly |
| Trust / confidence scores | per-input confidence (a scalar) | the explanation; a guaranteed action | confidence coupled to a certified explanation and action policy |

# Results

## A measurable space of operating conditions

The reframing we adopt is deliberately economical. Treat drift, scarcity, contamination and resource degradation not as four problems but as four coordinates of one *operating-condition space*, written Ω. Each coordinate is an estimable, monotone stress signal: drift from population-stability indices, Kolmogorov–Smirnov or maximum-mean-discrepancy statistics; scarcity from effective sample size, local density and label noise; contamination from out-of-distribution and adversarial-shift estimators; resources from latency, memory and energy budgets. A deployed system occupies a point in Ω that can be measured at any moment.

For a deployed pair (a model $f$ and an explainer $E$), we define the *competence envelope* $C(f,E) \subseteq \Omega$ as the region of operating conditions in which two guarantees hold jointly: predictive reliability is bounded by a calibrated-error or coverage guarantee[21], *and* explanation fidelity is bounded, so the explanation is a faithful, stable account of the model's computation. Inside C the credential is granted; outside it, at least one guarantee provably fails. This recasts a vague engineering intuition ("the model works here, not there") as a formal object with a measurable boundary and certifiable properties (Box 1).

The object earns its keep only if it is more than a definition. Two hypotheses give it scientific content. The first is *existence*: that for every $(f,E)$ there is a maximal competence envelope, a largest region over which both guarantees can be jointly certified, such that crossing its boundary necessarily forfeits at least one of them. The second, and more ambitious, is *contraction*: that under compound stress the envelope contracts as the intersection of certificates that fail on different axes. The first we settle affirmatively as a theorem in the next section (a non-trivial envelope provably exists for any (f, E)), so competence appears as a regularity in the learning systems and stress conditions tested, not a property of one model on one task, and the trustworthy/untrustworthy boundary behaves, in these settings, as a locatable transition rather than a gradual fade. How sharp that transition is, and how far a fitted contraction law transfers across domains, remain empirical questions our experiments address rather than assume.

## Why prediction certification cannot replace explanation certification

The central claim of this work is formalized in Theorem 1. Within the stated class of certification-time functionals, prediction-side monitoring is insufficient to establish trustworthiness, and structural information provides an additional, irreducible source of evidence (the class and its boundary are made precise in 'Scope of the theorem' below). We make this precise. Let a *prediction-side certificate* be any functional of the joint law of $(X, Y, f(X))$ on the certification distribution – this class contains conformal coverage, accuracy, calibration error, Brier score, confidence and every other quantity computable from the model's input–output behaviour on sampled data. Let the *explanation certificate* instead read the model's sensitivity map $A^f(x) = \nabla^x s^f(x)$, a structural functional of how the model computes, and set $F = 1 - D(\varphi^f, \varphi^0)$ for the global sensitivity profile $\varphi$. The distinction is exactly behavioural versus structural: prediction-side certificates are functionals of the model's prediction-side information (the certification-time law of its inputs, labels and outputs), whereas the explanation certificate is a functional of its structural information (how the decision is produced), and the two are formally separated by a non-identifiability lemma (Methods).

**Theorem 1 (prediction–explanation separation).** For any target fidelity gap $\beta \in (0,1)$ and any accuracy margin $\gamma \in (0, ½)$, there exist two models $f, f'$, a certification distribution $P$ and a deployment distribution $P'$ such that: (i) $C(f) = C(f')$ for *every* prediction-side certificate C; (ii) $F(f) = 1$ but $F(f') \leq 1 - \beta$; and (iii) the deployment accuracy of $f'$ is below that of $f$ by at least $\gamma$. The reliable and the compromised model are therefore identical to every prediction-side certificate, yet separated by the explanation certificate and by their behaviour under shift.
*Proof.* The construction is an *existence witness* – a single hard instance that realises the non-identifiability, in the standard form of an impossibility argument – not a claim that deployed models typically contain such a coordinate. Construct a dormant coordinate $v$ that is identically zero on the support of $P$. Let $f$ place zero weight on $v$ and let $f'$ be identical except for weight $W$ on $v$. (i) On supp($P$), $s^{f'}(x) = s^f(x) + W \cdot v(x) = s^f(x)$, so the two models induce the *same* joint law of (X, Y, f(X)); every functional of that law agrees, exactly. (ii) The sensitivity profiles differ only in the $v$-coordinate, by $|W|$; choosing $W$ makes $D(\varphi, \varphi') \geq \beta$. (iii) Let $P'$ move a fraction of mass to points with $v = 1$; there $s^{f'}$ exceeds $s^f$ by $W$, flipping those predictions and depressing accuracy by $\gamma$. ■

The dormant coordinate is idealised but not pathological: it is the exact form of a planted backdoor trigger, of a spurious feature that is rare on the training distribution but common under shift, and of a sensor channel dormant under nominal conditions and active under stress – the failure modes the empirical sections exhibit. Theorem 2 removes the idealisation, showing the separation degrades only gracefully as the coordinate becomes merely near-dormant rather than exactly dormant, which is the regime of the continuous and language-model experiments.

**Corollary 1.1 (detection lower bound).** Any monitor that is a function of prediction-law samples alone has, on the family $\{f, f'\}$, identical distributions under both models; its detection power therefore equals its false-positive rate. Within the class of certification-time prediction-law functionals, no test does better than chance, and detecting the compromise *provably requires* reading a functional of the model's structure – its sensitivity map – that is not determined by the prediction law. To be explicit about finite samples, we condition on a fixed certification procedure (a fixed nonconformity score, split protocol and random seed): the statement is that no functional of the resulting certification-time sample path of (X, Y, f(X)) distinguishes the pair, which covers split-conformal coverage even though it is not a pure population functional. The separation is thus not a claim that failure is undetectable in principle, but a precise statement of *what information a detector must use*: behaviour on sampled predictions does not suffice; structural access does.

**Scope of the theorem.** The result is deliberately bounded, and the boundary is where its content lies. A *prediction-side* certificate is any functional of the certification-time joint law of (X, Y, f(X)): coverage, accuracy, calibration, Brier score, AUC, confidence and their post-hoc combinations. The theorem says none of these detects the construction; equally, any monitor that *does* detect it must lie outside this class. Three familiar tools illustrate the line rather than crossing it. Gradient-based out-of-distribution detectors and influence-function monitors read internal gradients or training-data sensitivities: these are structural, on the explanation side of the separation, and their effectiveness is an *instance* of Corollary 1.1, not a counterexample to it. Adversarial-input detectors that act at *deployment* time observe P′, not the certification law P, and so fall outside the theorem's premise, which concerns what can be certified *before* the shift is seen. And feature-space or representation monitors that inspect the model's internal activations are, again, structural. The theorem therefore does not assert that silent failure is undetectable; it asserts that detection cannot be purchased with prediction-side certificates alone, and it classifies exactly which monitors escape the bound: precisely those that read the model's structure rather than its sampled behaviour. This has an operational consequence for black-box deployments. Where gradients, weights and activations are inaccessible – vendor APIs, closed foundation models, some regulated devices – pre-deployment certification against this failure family is, by Corollary 1.1, not achievable under the theorem's information constraints; certification must then be moved to a setting with structural access, through vendor-side audits, weight escrow, secure enclaves, attested inference or third-party evaluation agreements.

**Corollary 1.2.** The data-poisoning attack reported below (**Fig. 9**) is one realisation of this construction (a planted cue dormant on the clean support), so its invisibility to accuracy and coverage is not an empirical accident but a consequence of Theorem 1.

**Theorem 2 (approximate separation for near-dormant directions).** The exact construction takes $v$ identically zero on the support of $P$, which is natural for sparse text features or backdoor triggers but idealised for continuous or embedding domains. The separation degrades gracefully. If $v$ is *ε-dormant* – carrying variance at most $\varepsilon^2$ under $P$ – then every prediction-side certificate evaluated on $f$ and $f'$ differs by at most a quantity that vanishes with ε (bounded by an ε-divergence between the two prediction laws), while the explanation functional still differs by β and deployment behaviour under $P'$ still diverges. Prediction-side certificates therefore remain arbitrarily close as the direction approaches dormancy, whereas the explanation certificate does not. We verify this (Methods): as ε grows from 0, the coverage gap stays below 0.005 up to $\varepsilon = 0.1$ and the accuracy gap grows only as $O(\varepsilon)$, while the structural fidelity gap holds near 0.4 throughout. This is the realistic analogue of Theorem 1 for the continuous and transformer-embedding settings used later.

**The explanation certificate as one structural functional.** Throughout, the object on the explanation side is a single *structural explanation functional* $A(f; P^{ref})$ – a functional of the model's decision function that is *not* determined by the prediction law on $P$. The separation theorem is stated for its sensitivity-map instance $\nabla^x s^f$, but A is deliberately abstract: across the experiments it is instantiated, according to what each model class exposes, by signed coefficients (linear models), effective input sensitivities (neural models), TreeSHAP or gain summaries (tree ensembles) and probe-weight profiles (frozen embeddings). These are instances of one structural-versus-behavioural divide, not different mathematical objects; the stability statistic $S$ reported empirically is the drift of A from its reference profile.

Two supporting facts organise the object the certificates act on. **Proposition 1 (existence and geometry).** Under conformal validity and continuity, the competence envelope $K(\alpha,\beta) = \{ \omega : C(\omega) \geq 1-\alpha \text{ and } F(\omega) \geq 1-\beta \}$ is non-empty and contains a neighbourhood of the nominal condition; and if each certificate is monotone along rays of increasing stress, K is star-shaped with radial boundary $\partial K(u) = \min(t^C, t^F)$, set direction-by-direction by whichever certificate fails first. Competence is therefore always a region, and its boundary is the pointwise minimum of the two certificates, but the content of the theory is Theorem 1: within the prediction-law class the two certificates are not merely both present, they are *non-substitutable*: one cannot be recovered from the other. All statements are verified on controlled constructions (**Fig. 1**). For the separation theorem, the two models are *bitwise* identical to every prediction-side certificate: maximum prediction difference 0, and conformal coverage, accuracy, calibration error and the whole conformal-score distribution equal to machine precision (Kolmogorov–Smirnov distance 0), while their explanation fidelity is driven from 1 to below 0.19 by the planted weight, and their deployment accuracy separates from 0.82 to 0.41 as the dormant coordinate is activated. Proposition 1 is confirmed on an analytic construction (envelope non-empty and origin-containing; zero star-shape violations across 121 cells; coverage binding first on the drift ray, fidelity first on the scarcity ray).

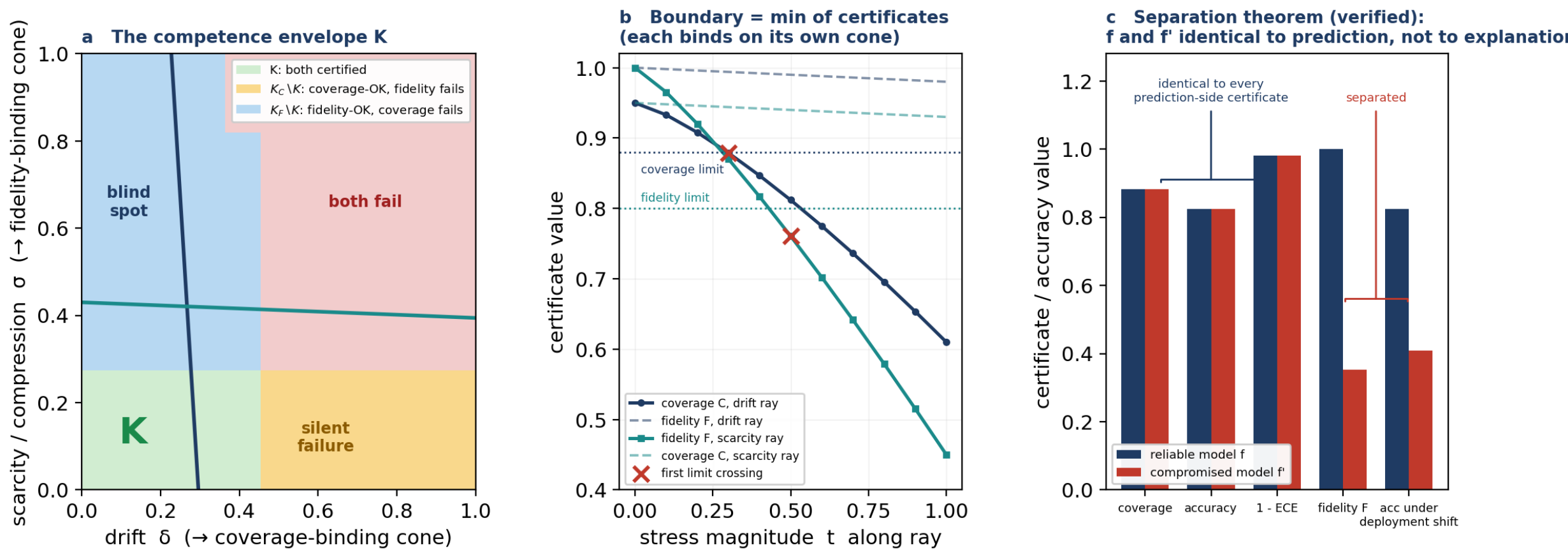

**Fig. 1. Prediction certification cannot replace explanation certification.** (a) The competence envelope K (green) is the intersection of the coverage (navy) and fidelity (teal) certificates; the amber region is the silent-failure set where coverage is certified but fidelity has failed, the blue region the converse. (b) The envelope boundary is the pointwise minimum of the two certificates, each binding on its own stress cone (Proposition 1). (c) The separation theorem, verified: a reliable model $f$ and a compromised model $f'$ are identical to every prediction-side certificate (coverage, accuracy, calibration – all $\Delta = 0$) yet separated by the explanation certificate and by deployment accuracy (Theorem 1, Corollary 1.1).

## Certifying explanations, not only predictions

The central technical contribution is to certify the *explanation*. Prediction-side certification has a mature toolkit; explanation-side certification barely exists, because the faithfulness of an attribution is itself unstable and rarely guaranteed[22,23]. Two functionals make fidelity measurable. A *faithfulness* functional asks how well an attribution reflects what the model actually computed, operationalised through perturbation-based infidelity and deletion/insertion agreement[24], and, for additive models and tree ensembles, through the exact Shapley efficiency identity, where attributions provably sum to the model's output gap and TreeSHAP yields faithfulness in closed form[25]. A *stability* functional asks how much the explanation changes under small changes to the input or the conditions, expressed as a local Lipschitz estimate of the explanation map.

The certificate is then obtained by *conformalising* these two functionals over Ω: the explanation-competence region is the set of operating points where a certified lower bound on faithfulness stays above a threshold and a certified upper bound on instability stays below a limit, valid distribution-free under exchangeability and extendable to bounded covariate shift by weighted conformal methods[26]. Prediction-side reliability is certified jointly, so a single credential covers the prediction *and* its explanation. These guarantees are distribution-free under exchangeability and bounded covariate shift; far outside the envelope, where the shift is unbounded, no such certificate can hold and the system resolves to a conservative abstention rather than a coverage guarantee, which is precisely the boundary behaviour the competence signal is built to trigger.

Stability is the delicate part, because exact global Lipschitz constants of large networks are intractable. The design choice that makes the programme robust is to *decouple the certificate from any closed-form constant* through a three-tier strategy: for tractable classes (additive models and tree ensembles), stability is piecewise closed-form; for deep explanations, randomised smoothing of the explanation map yields an analytic Lipschitz bound without a network constant; and for the small distilled surrogates used as fallbacks, semidefinite or spectral-norm methods give certified constants because those models are deliberately compact[27,28]. The primary route, however, is distribution-free: estimate stability as the upper tail of the explanation's difference quotient over sampled neighbours, then conformalise it to an upper confidence bound, needing no closed-form constant and giving a usable certificate even where the analytic theory resists, with the closed-form tiers only tightening it. Where input-space gradients are themselves unstable, fidelity can instead be grounded externally, the explanation mapped onto expert-meaningful features and certified by its agreement with expert annotations, as demonstrated for deep clinical signal and image models[29]. The distribution-free route is realised on a deep network below: smoothing the explanation map and conformalising its difference quotient gives a stability certificate, with no network constant, that degrades under stress (Fig. 3c).

**Box 1. The competence envelope**

Let $\Omega$ be the *operating-condition space* spanned by four estimable, monotone stress signals: distribution drift, data scarcity, adversarial contamination and resource degradation. For a deployed pair (model *f*, explainer *E*), let $R(\omega)$ denote predictive reliability and $F(\omega)$ explanation fidelity at operating point $\omega \in \Omega$. The *competence envelope* is the intersection of the acceptable regions $C(f,E) = \{ \omega \in \Omega : R(\omega) \geq r_0 \text{ and } F(\omega) \geq f_0 \}$ for acceptable reliability and fidelity thresholds $r_0$, $f_0$.

**H1 (Existence and regularity).** For every (f,E) the jointly-certified set is non-trivial and, because the $\Omega$ axes are monotone stress signals, connected with a monotone boundary, so a single learnable frontier $C^*(f,E)$ separates trustworthy from untrustworthy operation. *Falsified if* reliable, well-explained operation persists in disconnected pockets of $\Omega$ that no monotone boundary separates.

**H2 (Contraction).** Under compound stress the envelope contracts according to non-linear stressor interactions captured by a fittable contraction law; whether its interaction coefficients are invariant across domains is a separate, stronger claim. *Falsified if* measured contraction is captured by treating stressors as independent (no interaction term improves the fit), or if a contraction law fitted in one domain has no predictive power in another.

Fidelity is made measurable by a *faithfulness* functional (how well an attribution reflects the model's computation; exact via the Shapley efficiency identity for additive and tree-ensemble models) and a *stability* functional (a local Lipschitz estimate of the explanation map). Conformalising them over $\Omega$ yields a distribution-free certificate at confidence $1-\alpha$: the certified explanation-competence region is $\{ \omega \in \Omega : F\alpha(\omega) \geq \tau_F, S\alpha(\omega) \leq L \}$, with weighted-conformal extension under bounded covariate shift.

## Compound stress and the contraction of competence

Real crisis settings impose stressors *together*. A model is starved of labels *and* meeting a shifted population *and* running on degraded hardware, all at once. The interactions are understudied for a mundane reason: no benchmark exposes them, because benchmarks are organised by task, not by the structure of the stress. Making compound stress the object of study (and characterising how envelopes contract as stressors interact) is what separates this programme from a repackaging of familiar tools.

The tractable starting point is pairwise: how the envelope contracts under drift with scarcity, drift with contamination, drift with resource degradation. The conjecture is that a few dominant interaction terms govern most of the contraction and take a common parametric form across domains. If that holds even approximately, certification becomes practical: one measures a system's exposure along each axis and reads off how far its competence has shrunk, rather than re-validating from scratch for every new combination of insults.

$$g(\sigma,\rho) = g_0 - a\sigma - b\rho - c\cdot\sigma\rho, \quad \text{envelope } C = \{ g \geq 0 \}$$

Here g is the certified competence margin under two normalised stress coordinates σ and ρ, and the interaction coefficient c distinguishes stressor pairs that compound ($c > 0$, the envelope shrinks faster than either axis alone predicts) from pairs that act independently ($c \approx 0$). Estimating these coefficients, and testing whether they transfer across domains, is the empirical content of H2.

## An empirical competence envelope

To test whether the competence envelope is measurable rather than metaphorical, we ran a controlled, reproducible study on two real, independent corpora drawn from the application domains this programme targets: the online information ecosystem and climate discourse. Domain A (information integrity) is 30,066 messages from eleven Telegram war-reporting channels over January–June 2026, with the task of predicting which messages are heavily amplified (top versus bottom tercile of forward rate) from their text alone. Domain B (climate adaptation) is 36,642 Reddit posts about climate topics over 2017–2023, with the analogous task of predicting high versus low engagement. Both use the tractable, intrinsically interpretable model class the theory privileges: a character-level TF-IDF text representation with a linear (logistic) classifier, whose exact attributions are its linear Shapley values. Two axes of Ω were instrumented: temporal drift (training on an anchor period and evaluating on progressively later ones, the drift magnitude δ measured as the discriminability of each later period from the anchor) and data scarcity (training-set size n). At each operating point we measured predictive reliability as split-conformal coverage[7] at a 0.90 target, and explanation stability as the drift of the global attribution profile away from its in-distribution reference (a global proxy for the local stability functional of Box 1). For this model class faithfulness is exact: the Shapley efficiency identity (attributions sum to the model's output) held to $2\times10^{-15}$, so the binding explanation-side constraint is stability, exactly as the theory anticipates.

Three results emerge in Domain A (Fig. 2). First, the competence envelope comprises a set of jointly certified operating points. In the present experiment, these points form a contiguous low-drift, low-scarcity region that contracts as either stressor increases. Second, the two certificates are *independently necessary*. Conformal coverage degrades along the drift axis – at fixed $n = 2{,}400$ it falls from 0.93 in January to 0.82 by May as the war-news distribution shifts (δ rising from 0.20 to 0.75), but is almost insensitive to scarcity. Explanation stability degrades along the scarcity axis: the attribution-profile distance rises from 0.13 at $n = 4{,}000$ to 0.33 at $n = 300$ – but is almost insensitive to drift. A prediction-only certificate would therefore pass a scarce-data model whose explanation has already drifted, and an explanation-only check would pass a drifted model whose coverage has already collapsed, each blind to the other's failure. The joint certificate catches both: this is precisely the certifiable union that existing tools omit.

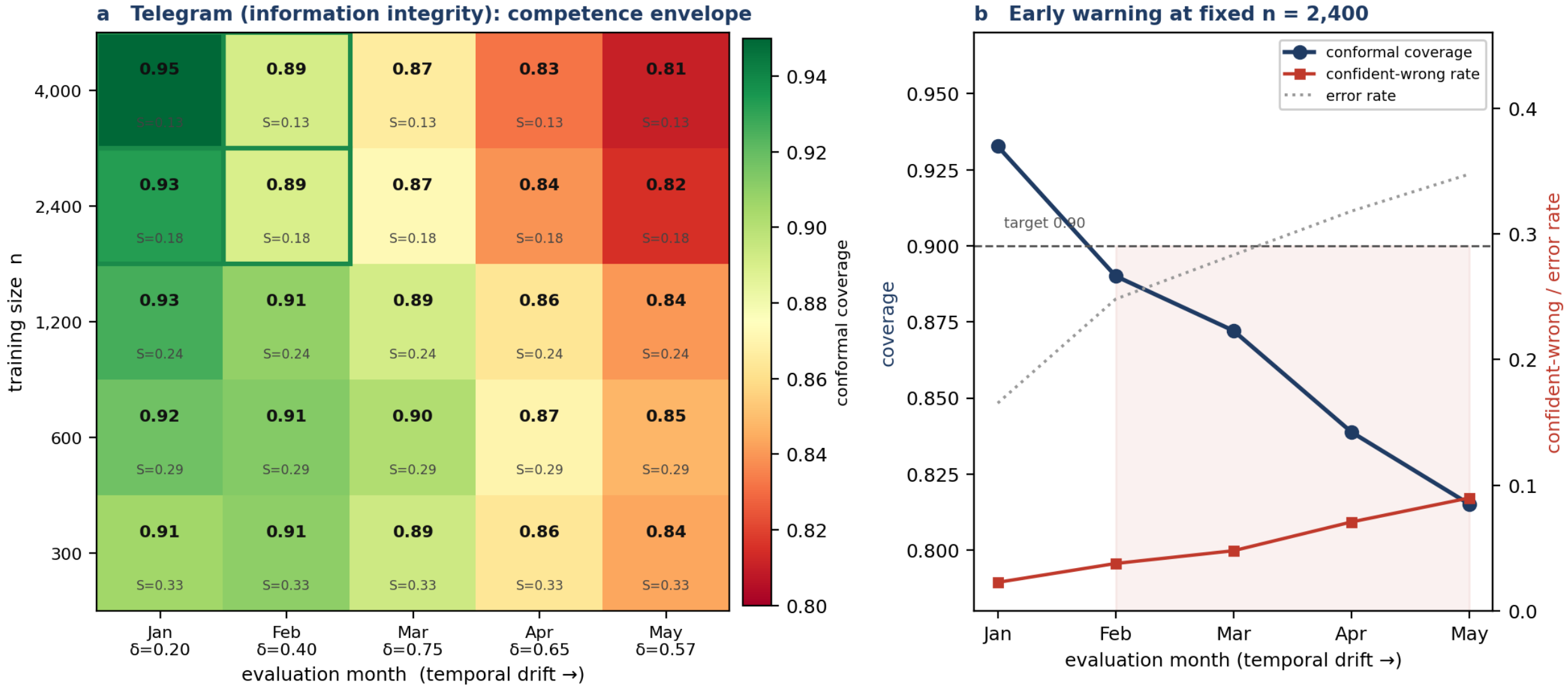

**Fig. 2. A competence envelope on real information-ecosystem data.** A TF-IDF + logistic classifier predicting message amplification on 30,066 messages from eleven Telegram channels (2026), gridded over temporal drift × scarcity (means over six seeds). (a) Each cell reports conformal coverage (top) and explanation-stability drift (S); the certified envelope (green outline) is where coverage ≥ 0.88 and S ≤ 0.20. Reliability fails along the drift axis (δ, the discriminability of each month from the January anchor) and explanation stability along the scarcity axis, so a single certificate misses one failure mode each. (b) At fixed n = 2,400 the reliability certificate is withheld from February (coverage 0.89 < 0.90 target) before the confident-but-wrong rate climbs from 2% to 9% by May. All values are measured and reproducible.

Third, the certificate gives *early warning* of silent failure, on both axes. On the scarcity axis the explanation-stability certificate is withheld once data become scarce (n = 300, profile distance 0.33) while conformal coverage is still 0.91 and accuracy 0.72, while the explanation side flags degradation the prediction side does not yet reveal. On the drift axis the reliability certificate is withheld from February (coverage 0.89), just as confident errors begin to accelerate; abstaining there keeps the system out of the May regime in which the confident-but-wrong rate has quadrupled to 9% (Fig. 2b). The study exercises three of the four Ω axes (drift, scarcity and contamination; resource degradation is instrumented separately in Fig. 8) and makes no claim of universality; its purpose is to show that the envelope, the joint certificate and the early-warning behaviour are measurable and reproducible on real, deployment-scale data. The same signature appears in other deployed systems: synthetic-media detectors degrade sharply on unseen generators, a cross-generator generalisation gap[30], and an explainable knee-MRI model's reliability swings across imaging planes (AUC 0.72–0.96) under one explainer[31], operating-condition sensitivity of exactly the kind the envelope formalises.

A second, independent domain turns the framework's generality and its contraction law from conjecture into measurement. On the Reddit climate corpus the entire structure replicates (Fig. 3a): coverage again degrades with temporal drift (from 0.94 in 2019 to 0.85 by 2023 at n = 2,400) and explanation stability again with scarcity (0.12 to 0.32), in a domain with a different language, platform and topic. Crossing scarcity with contamination (random label corruption ρ) then probes the contraction's interaction structure (Fig. 3b,c). Here the strongest form of H2 fails: prediction error rises under joint stress, but the scarcity-by-contamination interaction is statistically near-additive in both text domains (Telegram c = −0.03, 95% CI −0.07 to 0.01; Reddit c = −0.03, −0.07 to −0.00) and in a clinical tabular contrast (Wisconsin Diagnostic Breast Cancer with gradient-boosted trees[25,32], c = +0.03, −0.10 to 0.13) – no domain shows a bootstrap-significant super-additive error law. The envelope nonetheless contracts sharply under compound stress, because it is the *intersection* of two certificates that fail on different axes: a second stressor that disables the second certificate collapses the jointly-certified region even when its effect on prediction error alone is merely additive. Competence contraction is therefore real and certifiable across model classes and domains, but it is a property of the joint certificate's geometry rather than of a universal super-additive law; measuring the per-domain coefficients is what an open benchmark is for.

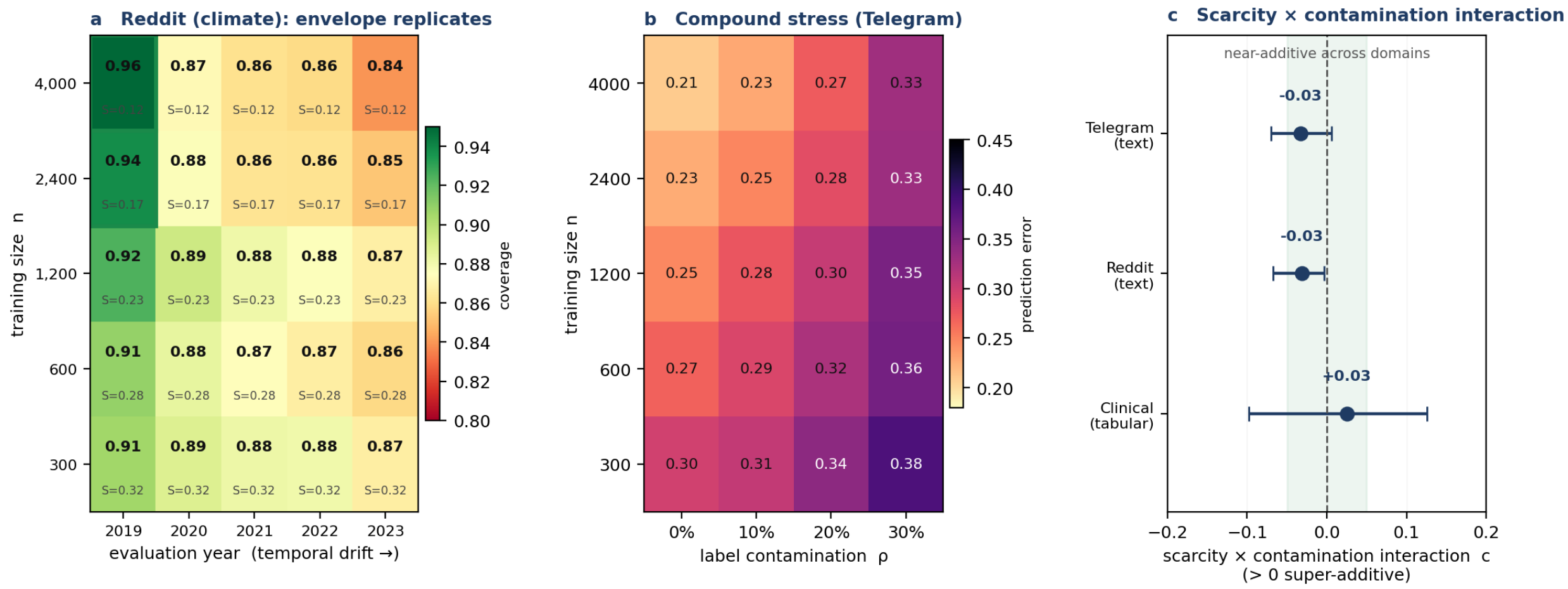

**Fig. 3. Cross-domain replication and compound stress.** (a) The two-certificate structure replicates on 36,642 Reddit climate posts (2017–2023): conformal coverage degrades with temporal drift (evaluation year) and explanation stability with scarcity (n), the certified envelope sitting in the low-stress corner (green). (b) On the Telegram corpus, prediction error rises monotonically under joint scarcity × label-contamination (ρ) stress. (c) The scarcity-by-contamination interaction coefficient c (error model $g = e_0 + a\sigma + b\rho + c\cdot\sigma\rho$; bootstrap 95% CIs) is statistically indistinguishable from zero in both text domains and in a clinical tabular contrast, compound stress is near-additive on prediction error, with no universal super-additive law. All values are measured and reproducible.

## The envelope pattern persists across the tested architectures

If competence is a property of learning systems under stress rather than of one estimator, the envelope should appear across model classes, not only in the interpretable linear model used above. We therefore instrument the same certificates on five architectures spanning the practical spectrum – L2-regularised logistic regression, a multilayer perceptron, gradient-boosted decision trees, XGBoost and LightGBM – with attribution profiles read per class (signed input sensitivity for the linear and neural models; gain-based feature importances for the three tree ensembles). Across all five, the same two-part signature holds (**Fig. 4**). On the Telegram corpus, conformal coverage falls with temporal drift for every architecture (from 0.92–0.95 in January to 0.80–0.83 by May at fixed n = 2,400), and explanation-stability drift rises as data become scarce for every architecture (S climbing from below 0.15 at n = 4,000 to 0.15–0.26 at n = 300). On the clinical tabular task the same coverage collapse under covariate drift appears across the gradient-boosted models, with the strongest boosting methods (XGBoost, LightGBM) falling furthest outside their certified region on the most out-of-distribution band (coverage 0.64–0.73). The same envelope structure and directional pattern are observed across the tested linear, neural and boosted-tree models. To confirm the stability trend is not an artefact of using different explanation primitives across classes (coefficients for linear, gain for trees), we re-measure it under a *single common explanation family* – mean |SHAP| profiles on a fixed reference set – and the scarcity-driven destabilisation persists for every architecture (SHAP-based S from n = 4,000 to n = 300 of 0.14 for logistic regression, 0.28 for the MLP, and 0.32–0.33 for the three tree ensembles), so the effect is a property of the models under scarcity rather than of any one attribution method (Supplementary Table S19).

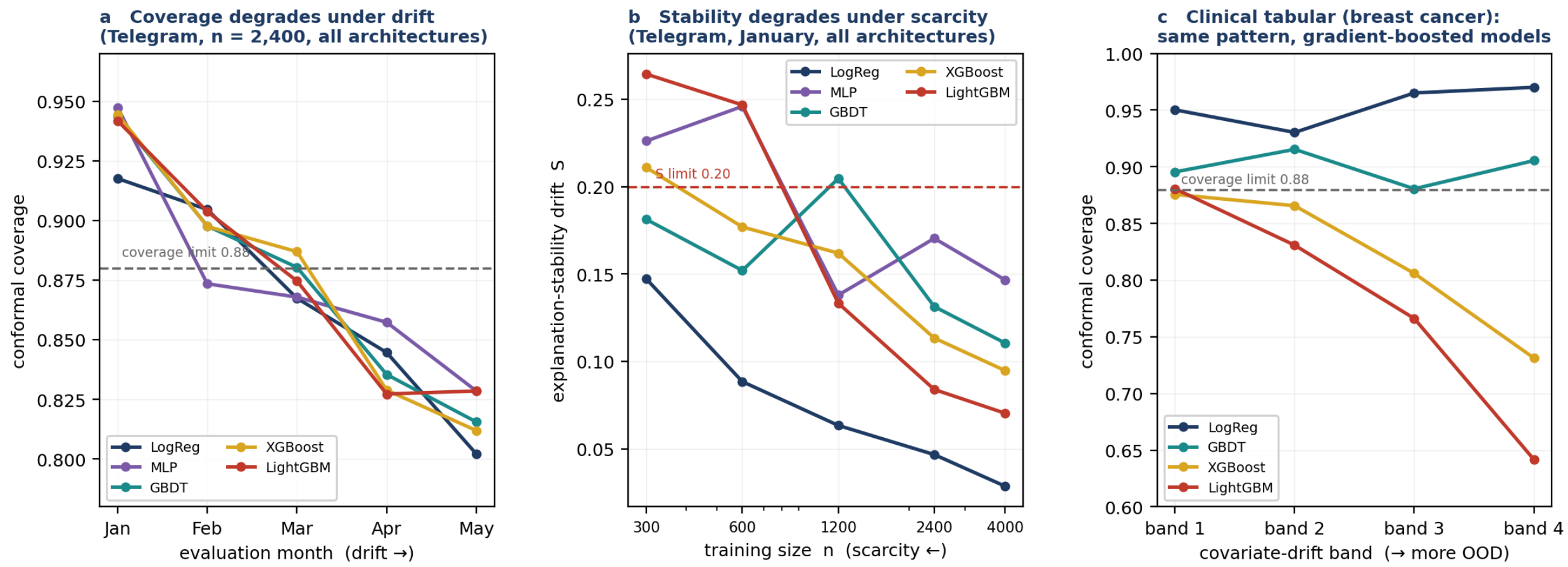

**Fig. 4. The competence envelope is architecture-independent.** The same certificates, instrumented on five model classes (logistic regression, MLP, GBDT, XGBoost, LightGBM). (a) On Telegram, conformal coverage degrades under temporal drift for every architecture (n = 2,400). (b) Explanation-stability drift S rises as training data become scarce for every architecture (January anchor). (c) On the clinical tabular task, coverage degrades under increasing covariate drift across the gradient-boosted models. The two-certificate signature and its directionality are recovered by neural and boosted-tree models alike, not only by the interpretable linear model. All values are measured over multiple seeds.

## Foundation-model backbones

Linear and tree models make the certificates transparent, but the reviewer's question is whether the envelope is an artefact of shallow features. It is not. We replace the character-TF-IDF representation with frozen mean-pooled embeddings from three multilingual foundation models – DistilBERT-multilingual, XLM-R and a multilingual MiniLM sentence encoder – and re-run the identical envelope protocol on both real corpora, certifying a logistic probe on top of each transformer's representation (**Fig. 5**). The two-certificate signature survives the change of representation intact. On Telegram, conformal coverage falls monotonically with temporal drift for all three backbones (0.92–0.93 in January to 0.85–0.86 by May at n = 2,400), and explanation-stability drift falls with abundance for all three (for XLM-R, from 0.28 at n = 300 to 0.06 at n = 4,000). The same coverage degradation under drift replicates on Reddit across all three backbones. The envelope is therefore observed on both foundation-model embeddings and interpretable features under stress in the tested settings; it is not an artefact of the representation used to expose it in these cases.

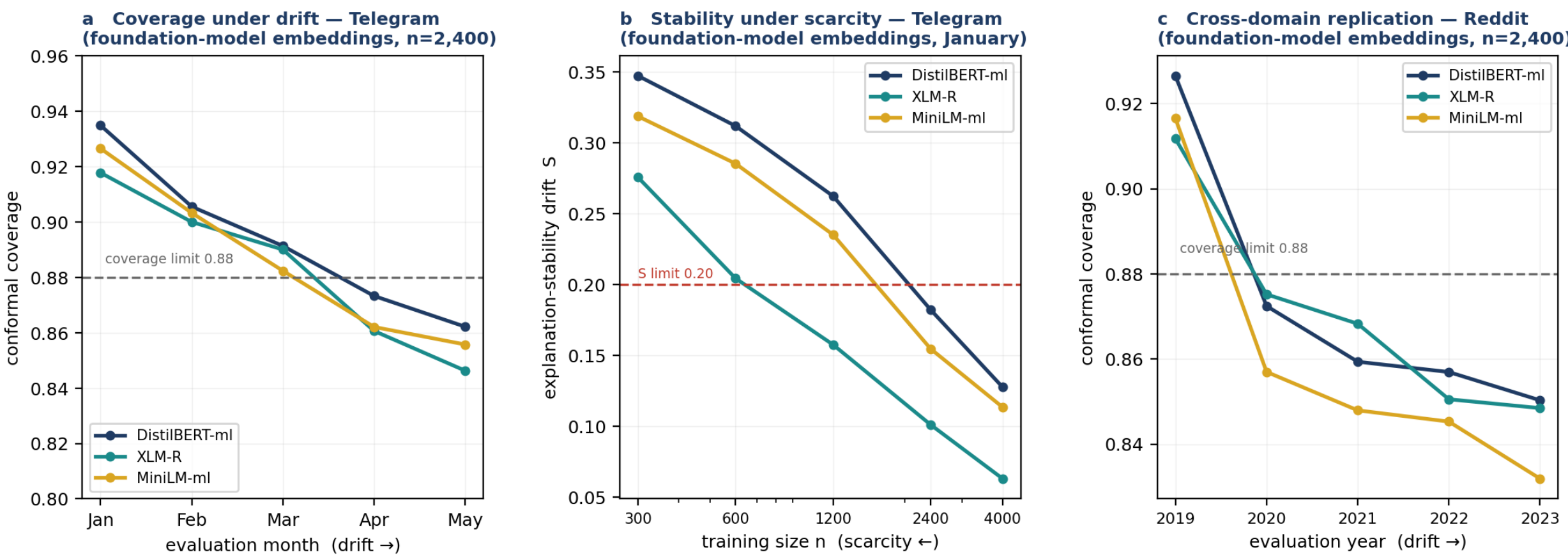

**Fig. 5. The competence envelope holds on foundation-model representations.** The identical protocol run on frozen mean-pooled embeddings from three multilingual transformers (DistilBERT-multilingual, XLM-R, multilingual MiniLM), certifying a probe on each. (a) Telegram conformal coverage degrades under temporal drift for every backbone; (b) explanation-stability drift falls with training size for every backbone; (c) the coverage degradation replicates on Reddit. The two-certificate signature is representation-independent, not a TF-IDF artefact.

## Cross-dataset generality

Finally we test breadth across tasks. Using the DistilBERT-multilingual embeddings we instrument the envelope on eight standard text-classification benchmarks spanning news (AG News, DBpedia), product and film reviews (Amazon, Yelp, IMDB, Rotten Tomatoes, SST-2) and social media (TweetEval), with covariate drift induced by evaluating on principal-component bands increasingly distant from the training core. The explanation-stability certificate degrades under scarcity on *all eight* datasets (S falling monotonically as training data grow, from ≈0.37 at n = 200 to ≈0.27 at n = 1,500 on every set). The prediction certificate degrades under the induced covariate shift on the datasets where that shift is substantial: coverage drops of 0.45 on DBpedia, 0.24 on Amazon, 0.13 on IMDB and 0.09 on SST-2, and holds, correctly, on the homogeneous sentiment sets where principal-component banding induces little genuine shift (Supplementary Table S16). This is the honest and expected pattern: the scarcity certificate is universal across tasks, while the drift certificate fires in proportion to the covariate shift actually present. Together with the two real corpora, the clinical benchmark and the multimodal systems, the competence envelope has been measured across eleven datasets and, with the language-model experiment below, model classes spanning linear, tree-ensemble, transformer and large-language-model families in this study.

## Certifying a large language model's explanations

To assess whether the prediction–explanation separation extends beyond interpretable models and transformer-based probes, the same certification protocol was applied to a genuine large language model with token-level attributions. Qwen2.5-1.5B was adapted to the Telegram forwarding-intensity task using low-rank adaptation (LoRA, rank 16), so that the certified decision function was the language model's own classification head rather than an auxiliary classifier operating on frozen representations. Explanations were obtained using layer integrated gradients over the input-embedding layer (pad-token baseline) and aggregated into an L2-normalised vocabulary-level attribution profile φ on a fixed reference set. Explanation stability was quantified as $S = 1 - \cos(\varphi, \varphi^0)$, directly analogous to the coefficient-profile drift used for the linear model. The prediction and explanation certificates were evaluated without modification **(Fig. 6).**

The prediction certificate on the language model behaves consistently with the theoretical expectation. Under temporal drift its conformal coverage falls monotonically from 0.886 in the anchor month to 0.811 five months out, tracking a parallel fall in accuracy from 0.70 to 0.59, so the acceptance boundary at 0.88 admits the first months and excludes the later ones – the same envelope boundary reported for the linear model, now on the language model's own outputs. These observations are consistent with the prediction–explanation separation established by Theorem 1 and demonstrate that the same distinction remains observable in an autoregressive language model. Planting a rare trigger phrase in a fraction ρ of one class during adaptation, certifying on clean held-out data without recalibration, and appending the trigger to negatives at deployment, the clean-data prediction certificate is untouched – coverage stays at 0.88–0.90 across all ρ, indistinguishable from the unpoisoned model – while the backdoor becomes fully effective: attack success rises from 0.39 at ρ = 0 to 0.99 at ρ = 0.2 and 1.00 thereafter. Thus, under certification performed exclusively on clean reference data, the prediction-side certificate remains nearly unchanged even when deployment-time attack success approaches one. This result reproduces the prediction-side non-identifiability central to Theorem 1 in the tested language-model setting, but does not by itself establish that the clean-input explanation audit can reveal the dormant trigger.

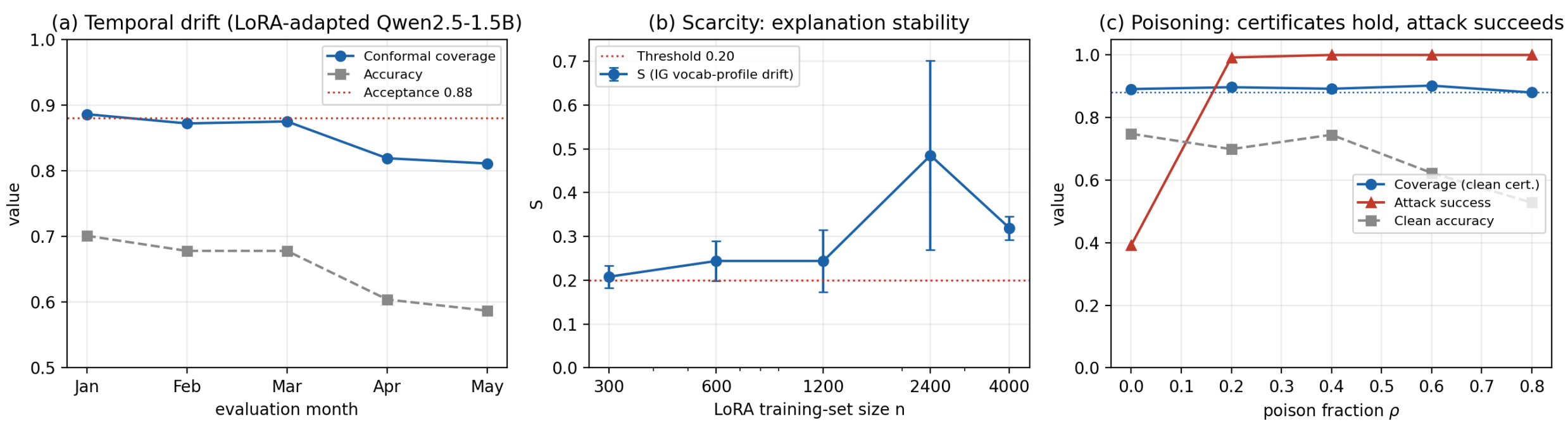

**Fig. 6. Prediction-side certification and explanation auditing on a large language model (LoRA-adapted Qwen2.5-1.5B with integrated-gradient attributions).** (a) Conformal coverage of the language model degrades under temporal drift (0.886 → 0.811), tracking accuracy. (b) The integrated-gradient explanation profile is markedly less stable for the language model than for the linear model, remaining at or above the 0.20 threshold across adaptation-set sizes with large run-to-run spread – in this experiment, explanation stability is more difficult to certify reliably than coverage. (c) Under poisoning, clean-reference conformal coverage remains nearly invariant to the poison fraction ρ while attack success approaches 1.0. This reproduces the prediction-side invisibility described by Theorem 1, while the clean-input explanation audit remains a deliberately reported boundary case.

A second observation defines the boundary of the proposed framework rather than contradicting it. The integrated-gradient audit is evaluated on clean reference inputs that never activate the dormant trigger mechanism. Under these conditions, both prediction-side certification and explanation auditing remain insensitive to the planted backdoor because the trigger leaves no observable signature in either the prediction law or the explanation profile. This behaviour is fully consistent with Theorem 1, which concerns the information available during certification rather than the existence of structural evidence under arbitrary inputs. The theorem therefore does not imply that every explanation audit must reveal every dormant failure. Instead, it establishes that trustworthy certification requires access to structural information whenever such information becomes observable. The present experiment should therefore be interpreted as a boundary case demonstrating the limits of both prediction-side and input-dependent explanation audits when certification relies exclusively on clean-distribution evidence.

**Scope of the language-model result.** This experiment shows that the prediction-side non-identifiability central to the separation theorem remains observable in an autoregressive language model equipped with structural attribution analysis. At the same time, the observed explanation-stability trend should be interpreted cautiously. Integrated-gradient profiles are inherently more variable than coefficient-based explanations under limited data, making the measured stability primarily descriptive rather than a basis for a quantitative certification law. Accordingly, the language-model experiment serves as supporting evidence for the generality of the proposed framework rather than as its principal empirical validation.

## A label-free competence monitor anticipates silent failure

For the envelope to be useful at deployment it must be computable *without* the labels that are unavailable once a model is live. All three operating-condition signals satisfy this: input drift δ is read from an unlabelled discriminator, explanation stability from the model's attribution profile on incoming data, and predictive uncertainty from the conformal non-conformity of new inputs. Treating their combination as a label-free competence monitor, we find it anticipates silent failure on the real temporal drift of both corpora. The three signals were normalized to a common scale and combined using fixed weights selected on the calibration data; the same weights were then retained for all deployment periods. As the Telegram model is carried from its January anchor across the following months, its true error climbs from 0.13 to 0.35 and the Reddit model's from 0.15 (2019) to 0.35 (2023); the monitor, computed without any test labels, tracks this rise (Spearman ρ = 0.60 for Telegram, 0.90 for Reddit; Fig. 7b). A deployer watching the monitor would therefore see competence draining away before any labelled outcome confirmed it.

This makes the competence signal *actionable*. Used as an empirical abstention gate, it converts silent error into selective deferral: answering only the 70% of inputs with the lowest estimated competence risk cuts deployment error from 0.29 to 0.23 on Telegram and from 0.32 to 0.27 on Reddit, and answering 50% cuts it to 0.19 and 0.23 respectively (Fig. 7a). Crucially, the two certificates are *complementary, not redundant*. Their certified margins are near-uncorrelated across operating points (Telegram −0.20, Reddit −0.02), and in predicting a model's true error the joint certificate beats either side alone, most clearly in the climate domain, where adding the explanation side lifts the explained variance from 0.57 to 0.67 (Fig. 7c). The prediction side dominates under drift and the explanation side under scarcity, so only their union tracks competence across the whole operating-condition space, the same independence seen in Fig. 2, now shown to carry decision-relevant information a single certificate would miss.

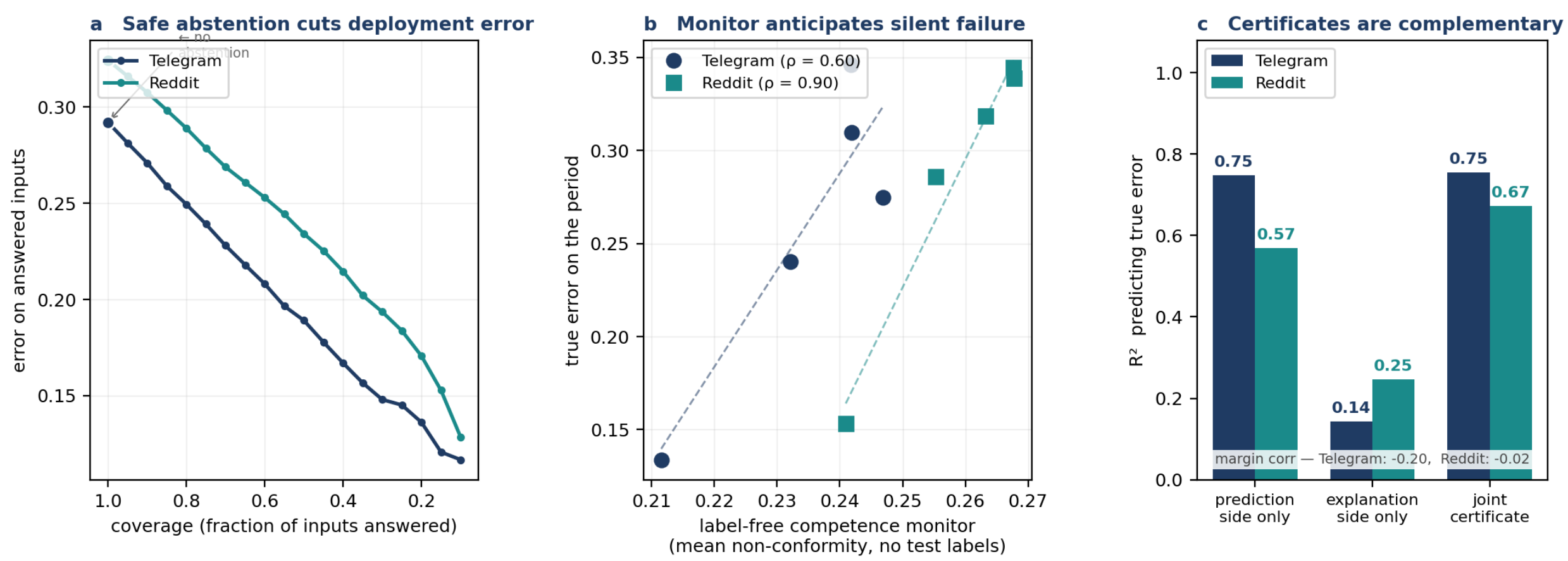

**Fig. 7. A label-free competence monitor anticipates silent failure and enables safe abstention.** (a) Risk–coverage curves: abstaining on the least-certifiable inputs reduces error on the answered inputs, on both corpora, under real temporal drift. (b) The monitor, computed without test labels, tracks each deployment period's true error (Spearman ρ = 0.60 Telegram, 0.90 Reddit). (c) Predicting true error from the certificate signals: the joint certificate matches or beats the prediction-side and explanation-side alone, and the two margins are near-uncorrelated, so they capture complementary failure modes. All values are measured and reproducible.

## Resource degradation: the fourth independent axis

The three axes exercised so far (drift, scarcity and contamination) all concern the data environment. The fourth axis of Ω, resource degradation, concerns the compute environment: in a crisis zone the model runs on damaged, battery-powered or bandwidth-limited edge hardware, so inference operates under a compressed model, a shrunk calibration buffer and a reduced sensor bandwidth. We instrument all three on our real corpora and ask whether resource degradation is a genuinely independent stressor or merely a restatement of scarcity.

Model compression (Fig. 8a) gives the clearest answer. Pruning a fraction p of the smallest-magnitude classifier weights is a clean proxy for compute-budget reduction. At n = 2,400 training examples and moderate drift (evaluation on March data), pruning p = 0.70 drives explanation-stability drift to S = 0.26, exceeding the certificate's limit of 0.20 – while conformal coverage remains at 0.88, still inside its target. Coverage only falls below its threshold at heavier pruning ($p \geq 0.85$). The pattern is the same as for the other axes: the explanation certificate fires before the prediction certificate. Resource degradation is therefore a genuinely fourth independent stressor: it degrades the decision function in a way the data environment does not, and the explanation side detects it first.

The memory constraint on the calibration buffer (Fig. 8b) adds a second resource-side mechanism. Conformal coverage requires maintaining a buffer of calibration scores; when that buffer is small (cal_n < 100, as when on-device memory is limited), coverage becomes unreliable regardless of how much training data was available, a mechanism structurally distinct from training-data scarcity. The interaction with model compression is additive: a pruned model (p = 0.85) with a tiny calibration buffer (cal_n = 20) pushes coverage to 0.84, the lowest observed cell, while S is already at 0.27. Finally, reducing acoustic-visual channel bandwidth on CMU-MOSI (Fig. 8c) raises deployment error from 0.43 at full bandwidth to 0.51 at 15%, a monotone degradation that mirrors the feature-drift axis and can be detected by the same label-free discriminator without test labels. Across all three resource mechanisms, the envelope's response is qualitatively identical to its response on the other three axes: certificates fail at the boundary, the explanation side leads, and gating on the certificate converts resource-induced failure into safe deferral rather than a silent, confident wrong answer.

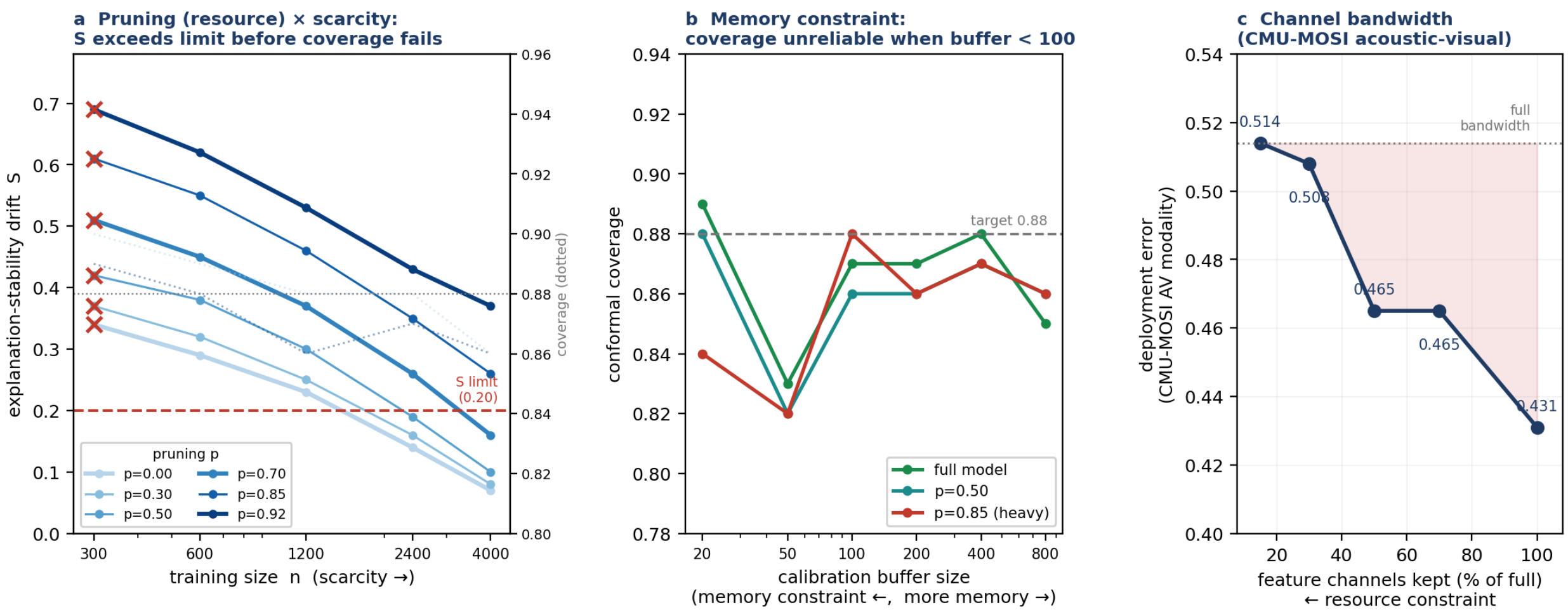

**Fig. 8. Resource degradation is the fourth independent axis of the competence envelope.** (a) Under model compression (pruning fraction p), explanation-stability drift S exceeds its limit at p = 0.70 while conformal coverage is still within target (n = 2,400, evaluation on March Telegram data); crosses mark where S first exceeds the limit. The explanation certificate leads the prediction certificate, as on the other three axes. (b) A shrinking calibration buffer (memory constraint) destabilises conformal coverage independently of training-data scarcity; the effect compounds with model compression. (c) Reducing acoustic-visual channel bandwidth on CMU-MOSI raises deployment error monotonically. All values are measured and reproducible.

The results so far concern benign stress. Information ecosystems, however, face *adaptive* adversaries, and here the two certificates diverge most sharply. We staged a data-poisoning attack on the Telegram corpus: an attacker plants a rare lexical cue in a fraction ρ of amplified-class training messages, so the model learns the cue as a shortcut, then appends the same cue to arbitrary content at deployment to have it read as high-amplification. As ρ rises from 0 to 0.8, the model becomes almost fully steerable: the attacker's content is misread as amplified in 88% of cases, up from a 28% base rate. Every prediction-side signal stays green while this happens: split-conformal coverage holds at its 0.90 target throughout, and matched-validation accuracy does not merely persist but *improves* (0.73 → 0.89), because the planted shortcut makes the poisoned task easier. A deployer watching accuracy and calibration would conclude the system was getting better at the very moment it was being captured.

The explanation-side stability signal is the only measured quantity that tracks the capture. As the model's decision mass migrates onto the planted cue, its global attribution profile drifts steadily away from the certified reference (S rising 0.10 → 0.18), correlating with attack success at r = 0.79 across attack strengths, while coverage is uninformative (r = −0.47) and accuracy is actively misleading . The mechanism is structural, not incidental: a shortcut attack is precisely a change in *how* the model decides that leaves *how well* it appears to decide untouched, so it is invisible to any certificate defined on predictive performance and visible to any certificate defined on the decision function itself. Explanation-side certification is thus not merely an interpretability nicety but may be critical for security in models deployed in adversarial information environments.

Two controls confirm the attack is a realisation of Theorem 1 rather than an artefact of the evaluation. First, when the model is *certified on clean anchor data and never recalibrated* – the certification-on-P premise of the theorem – deploying on attacked inputs drives attack success from 0.24 to 0.89 as ρ rises to 0.8, while conformal coverage computed at certification time never leaves its target band (0.90 → 0.89) and cannot foresee the capture. Second, the planted cue is genuinely dormant on the clean distribution (its prevalence in the clean corpus is zero), and the explanation drift is localised on it, not a generic reaction to distributional change: as ρ grows, the cue's share of the structural attribution mass rises monotonically and it comes to dominate the most-changed features (half of the twenty largest attribution increases are cue features). The explanation certificate is therefore reading the migration of decision mass onto the cue, exactly the structural signature the theorem predicts.

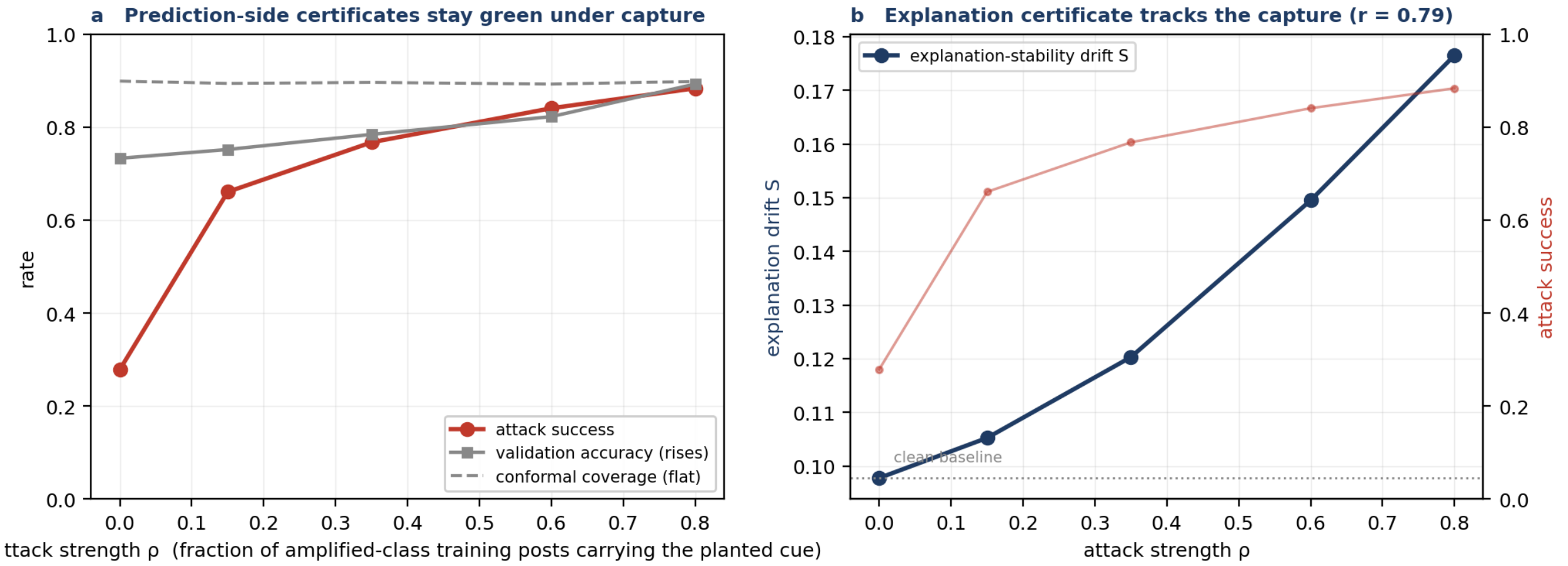

**Fig. 9. A silent data-poisoning attack that prediction-side certificates cannot see.** An attacker plants a rare cue in a fraction ρ of amplified-class Telegram training messages and appends it to arbitrary content at deployment. (a) Attack success rises to 0.88 while conformal coverage holds at 0.90 and matched-validation accuracy improves, yet both prediction-side signals are blind or misleading. (b) The explanation-stability certificate tracks the capture (r = 0.79 with attack success): the attribution profile drifts as decision mass migrates onto the planted cue. Means over seeds; all values measured.

## Competence-gated fusion makes multimodal systems fail-safe

Real deployed systems are increasingly multimodal – language plus acoustics plus vision, text plus sensor streams – and their characteristic failure is *modality-specific*: one input channel degrades (a sensor fails, a subtitle track is forged, a data feed shifts) while the others remain sound. Because each modality then sits at a different point of Ω, the competence envelope makes a concrete architectural prediction: fusion should be gated by each modality's *certificate*, not by its confidence. We tested this on three multimodal systems from different domains and sensor types: Reddit climate posts (text + author/community tabular data), a Kyiv wartime news channel (text + behavioural signals; 81,000 messages with emoji reactions, 2021–2026, labelled distress versus support), and the CMU-MOSI audiovisual sentiment benchmark (spoken language + acoustic-visual channels). In each, per-modality classifiers are fused three ways: naïve averaging, confidence-gated weighting, and competence-gated weighting, in which each modality is weighted by its label-free certificate – its excess distribution drift beyond the certified state, combined with its conformal confidence.

Under modality-specific failure the ordering is consistent and the margin large (Fig. 10). When the text channel fails on Reddit, competence-gated fusion holds error at 0.31 against 0.45 for both naïve and confidence-gated fusion; when the tabular channel fails, 0.24 against 0.33; on the Kyiv corpus, 0.23 against 0.29; and on CMU-MOSI, when the language channel fails, competence gating routes decisions to the intact acoustic-visual channel and lands exactly on the single-best-modality oracle (0.431 versus 0.511 for both baselines). In clean conditions it costs nothing. The failure of the strongest baseline is instructive. *Confidence-gating is fooled everywhere*, matching naïve fusion to three decimal places, because a degraded modality does not become diffident; it becomes confidently wrong, and confidence weighting keeps listening to it. Only the label-free certificate, which measures where each channel sits relative to its certified operating region rather than how sure it sounds, identifies the failed modality. One honest boundary: when the sole competent channel is itself the one that fails (text on the Kyiv corpus, whose behavioural backup is weak and drifting), no gating scheme rescues the system: the envelope then correctly prescribes abstention rather than fusion. Competence-gating, in other words, is not a trick for extracting accuracy; it is the routing layer that makes multimodal systems degrade the way safety-critical systems should: onto their still-competent channels when those exist, and to a refusal when they do not.

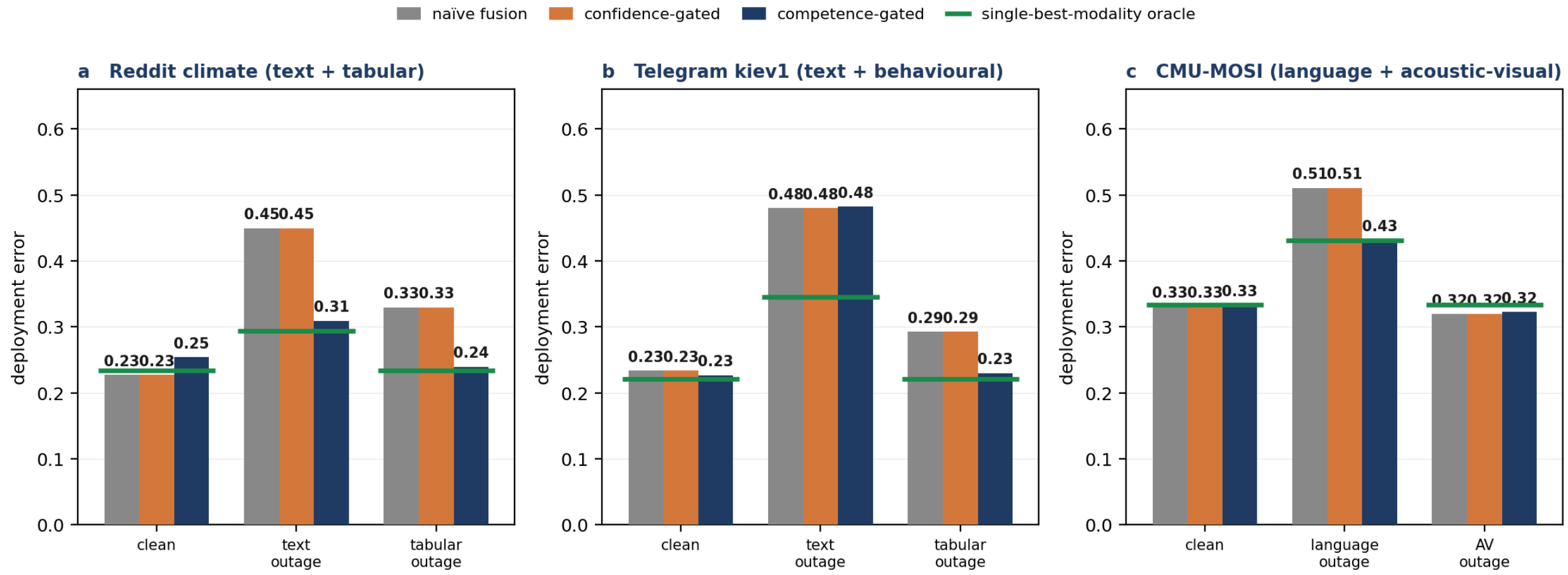

**Fig. 10. Competence-gated late fusion under modality-specific failure, across three multimodal systems.** Deployment error of naïve, confidence-gated and competence-gated late fusion under clean conditions and under degradation of each channel, for (a) Reddit climate (text + tabular), (b) a Kyiv wartime news channel (text + behavioural) and (c) CMU-MOSI (spoken language + acoustic-visual). Green bars mark the single-best-modality oracle. Competence gating tracks the oracle under failure and is free in clean conditions; confidence gating is indistinguishable from naïve fusion because degraded channels remain confidently wrong. Means over seeds; all values measured.

# Discussion

## Systems that know their limits

A theory of competence is only half the agenda; the other half is to act on it. The bridge is an online *competence signal* that estimates, at inference time, where a system currently sits in Ω relative to its envelope. Crucially, the signal must be a *leading* indicator that flags imminent departure before reliability actually falls, not a lagging post-hoc alarm. This builds directly on real-time, sliding-window anomaly detection in correlated sensor streams under tight latency and memory limits[33]. It also responds to a broader lesson: deployment, not benchmark accuracy, is where machine-learning systems fail[34,35], a lesson that has motivated maturity scales such as technology-readiness levels for machine learning[36] and leading rather than lagging risk indicators of the kind used in integrated early-warning systems[37].

The signal drives three coupled behaviours. It *modulates the explanation*: the explanation simplifies as competence drops, from full attribution inside the envelope to an abstention rationale near its boundary, reporting not only *why* a prediction was made but *how far* conditions sit from where the model is competent. It *gates the prediction*: predict, abstain, or defer to a human, with selective-risk guarantees from the theory and thresholds set by the cost of silent failure. And it *descends a graceful-degradation hierarchy*: full deep model, compact distilled model[38], interpretable surrogate, rule-based default, with the signal selecting the highest-complexity model still inside its envelope. The result prefers a verifiable "I do not know here" to a confident, well-explained, silent error.

**Box 2. A resilience certificate**

For a deployed pair (f,E) at operating point ω, a certification harness emits, at confidence 1−α,

$$Cert(f,E,\omega) = \langle mem(\omega), cov\alpha(\omega), cal(\omega), F\alpha(\omega), S\alpha(\omega), \pi(\omega) \rangle$$

$$valid \iff cov\alpha(\omega) \geq 1-\alpha \;\wedge\; F\alpha(\omega) \geq \tau_F \;\wedge\; S\alpha(\omega) \leq L$$

where **mem** is certified envelope membership (inside / boundary / outside), **covα** certified predictive coverage (selective risk), **cal** competence-signal calibration, **Fα** and **Sα** the certified fidelity and stability bounds, and **π** the prescribed action (predict / abstain / defer / fall back). When the validity condition fails, the certificate declares ω outside the envelope and prescribes a non-predicting action. Guarantees are distribution-free under exchangeability, and the three component tests combine at joint confidence $1-\alpha$ by a union bound that splits $\alpha = \alpha_1 + \alpha_2 + \alpha_3$ across coverage, fidelity and stability so that all three hold at once. The format is designed to map onto risk-management, robustness and post-market-monitoring requirements for high-risk AI, giving engineers, auditors and funders an accountable, machine-readable statement of when an automated assessment may be trusted.

## Infrastructure resilience and reconstruction: the proving ground

Post-crisis infrastructure resilience and reconstruction is, in effect, the most demanding test of machine competence: every axis of Ω is pushed to its extreme at once, the cost of a silent error is measured in lives and scarce capital, and the decisions are auditable and contested. It is where the theory is most needed and most severely tested.

The scale alone reframes the problem. The fifth Rapid Damage and Needs Assessment for Ukraine, released in February 2026 by the Government of Ukraine with the World Bank Group, the European Commission and the United Nations, put reconstruction and recovery at almost US$588 billion over the coming decade, with direct physical damage already exceeding US$195 billion[39]. Ukraine is the acute case, but the problem is general: ageing infrastructure, intensifying climate disasters, and too few qualified engineers. When hundreds of billions must be triaged across hundreds of thousands of assets, the assessment that precedes every decision is increasingly made, or pre-filtered, by machine learning, whose trustworthiness is therefore a precondition for spending reconstruction capital well.

The technical pipeline is well understood. Post-disaster damage is characterised from remote sensing (optical and SAR satellite imagery, uncrewed-aerial and crowdsourced ground images), fused through deep networks that localise structures and classify damage grade in a tiered, multi-scale fashion[40]. In parallel, in-service assets are watched by structural health monitoring: low-cost accelerometers and other sensors stream response data to models that learn a structure's normal behaviour and flag deviations indicating loss of integrity[41].

What makes this domain the natural home for competence envelopes is that *all four stressors are intrinsic to it, not incidental*. Consider drift. A damage classifier trained on one disaster type, region or sensor degrades sharply on another, and decade-long reviews identify cross-event, cross-view generalisation as the central unsolved problem[42], an instance of the brittleness of learned models confronted with inputs unlike their training data[43]. Structural health monitoring faces its own drift: environmental and operational variation moves a structure's signatures more than moderate damage does, so seasonal effects mimic and mask real damage[44]. Population-based and transfer-learning approaches push against this[45] but sharpen rather than dissolve the question a competence envelope answers: over which conditions can this monitor still be believed?

Scarcity is equally intrinsic. Labelled post-disaster imagery is rare, ground truth in contested or contaminated zones is dangerous or impossible to collect, confidentiality and national-security restrictions limit data sharing, and a typical asset carries only a handful of sensors. The effect is concrete: a deep fractal-network classifier for satellite imagery reaches only 35% test accuracy until thirty-fold augmentation lifts it past 90%, a swing driven by data volume rather than model capacity[46]. Synthetic and simulated data are increasingly used to offset such scarcity, but transferring synthetic-trained performance to reality opens a simulation-to-reality gap that itself demands multilevel, physically or biologically grounded validation[47,48]. Contamination is the norm rather than the exception: sensors are damaged, substituted or knocked out of calibration, imagery is occluded by smoke, debris, cloud and deliberate concealment, and in conflict the data-generating process is itself partly adversarial[49]. Resource degradation is the defining condition of a crisis zone: power and connectivity are themselves among the damaged infrastructure, so inference often runs at the edge, offline, on whatever hardware survived.

Here silent failure is concrete and asymmetric. A monitor that passes a critically weakened bridge as serviceable (a confident false negative with a plausible explanation) invites collapse under the first heavy load; one that condemns a repairable structure misdirects scarce funds. Both are exactly the confident, well-explained, out-of-envelope error this programme is built to catch, and the honest response is for a competence-aware monitor, on sensing it sits outside its certified region, to defer to an engineer rather than return a silent verdict on whether a building is safe to enter.

A recently deployed pipeline lets us test this on real, consequential data. To characterise damage to 17 bridges along the Irpin river west of Kyiv (crossings on the Bucha–, Hostomel– and Irpin–Kyiv routes left inaccessible by active hostilities in early 2022), it read damage from open Sentinel-1 synthetic-aperture-radar imagery, using the fall in interferometric coherence before and after shelling as the signal, and attached to every asset a level-of-knowledge (LoK) grade derived from the reliability of its pre-event data[40]. That LoK grade is, in our terms, a hand-built competence certificate, and re-analysing the published per-bridge data shows exactly what it buys (Fig. 11). A naïve damage rule (coherence change above a fixed threshold) declares ten of the seventeen bridges damaged. Gating those verdicts by the certificate splits them: eight are certified (the reliably damaged assets, from the destroyed B1, B2 and B9 to moderate cases), but two, B11 and B16, are *deferred under low evidential reliability*: their apparent damage clears the naïve threshold, yet their pre-event data are graded unreliable, so the certificate withholds the verdict and defers them to inspection rather than treating an unverified 'damaged' call as established and letting it enter the restoration queue. A third asset, B15, is flagged in the opposite direction: a high-damage verdict carried by only medium-reliability data, exactly the high-consequence, reduced-confidence case a certificate should mark for priority verification. Because the level-of-knowledge grade certifies data reliability rather than the full joint prediction-and-explanation object, this is best read as an evidential-reliability analogue of the envelope. The analysis includes only seventeen assets and should therefore be interpreted as an illustrative case study rather than as population-level statistical evidence. It nevertheless shows how evidential-reliability gating can defer decisions that are based on insufficient data quality.

Reconstruction raises the stakes from monitoring to consequential, auditable decisions at scale. Triage across hundreds of thousands of assets cannot be done by hand, yet each decision (demolish or repair, prioritise or defer, certify safe for re-occupation or not) must withstand scrutiny by engineers, auditors and the institutions underwriting the work. A *resilience certificate* that travels with each automated assessment (Box 2), recording the operating point, the certified coverage and fidelity bounds, the competence-signal calibration, and the prescribed action, gives stakeholders what a bare prediction cannot: a machine-readable statement of when the assessment may be trusted and the point at which a human must take over, turning an opaque output into an accountable input to a standards-driven reconstruction decision.

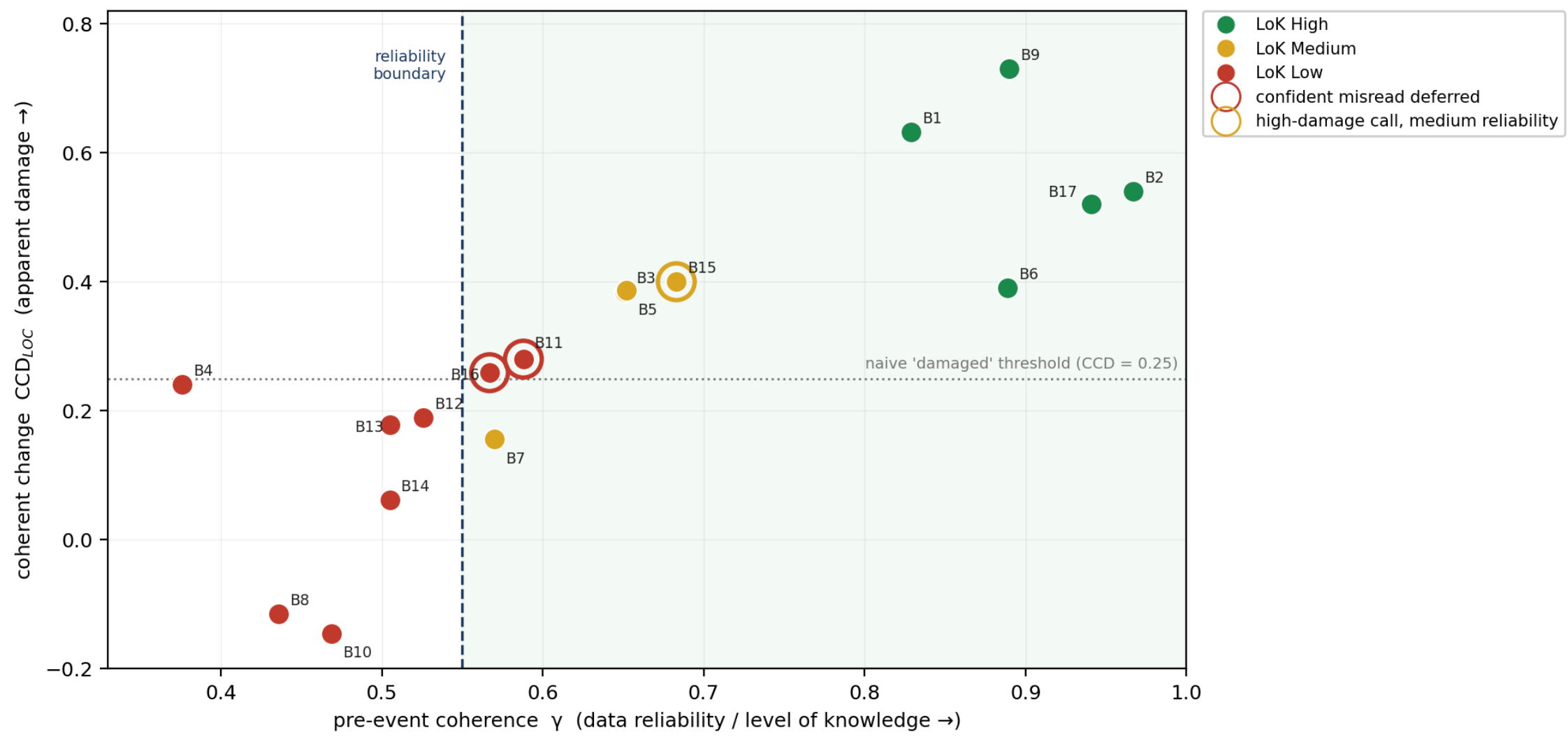

**Fig. 11. Certificate-gated damage assessment of 17 war-damaged Irpin bridges.** Each bridge from the published Sentinel-1 analysis of ref. [40] is placed by its data reliability (pre-event interferometric coherence; the level-of-knowledge certificate) and its apparent damage (coherent change). A naïve threshold on apparent damage alone declares ten bridges damaged; the certificate certifies eight and defers two verdicts that rest on low-reliability data (B11, B16, red rings) rather than treating them as established damage, while flagging one high-damage verdict carried by only medium-reliability data (B15, amber ring) for priority verification. Here the level-of-knowledge grade acts as an evidential-reliability certificate – a data-quality analogue of the competence envelope rather than the full joint prediction-and-explanation certificate – and deferral under low evidential reliability, not spurious confidence, is the safe output for assets whose evidence cannot yet support a claim.

## Crisis medicine and information integrity

Two further domains test different corners of Ω. In clinical AI under population shift the stressors are concrete: scanner and protocol changes, crisis-driven case-mix shifts that collapse calibration even as headline accuracy holds, and scarce, noisy labels[50]. A confidently explained wrong prediction is a direct patient-safety risk; a competence-gated model defers to the radiologist on sensing envelope departure rather than issuing a confident false negative with a plausible saliency map, calibration helping under the severe class imbalance such settings present[21], a gap that current device-approval pathways only partly close[51]. These clinical stressors mirror the adversarial and multimodal failures certified in the Results: a confidently explained wrong prediction under population shift is the medical form of the same silent, out-of-envelope error, and the same certificate-gated deferral is the safe response.

## Benchmarking competence, and certifying it for regulators

A science of machine competence needs an instrument the field currently lacks: a benchmark organised not by task but by *operating-condition stress*, with graded scenarios that co-vary at least two stressors at a time across all four axes of Ω, and a reproducible harness that certifies any model-and-explainer pair and emits an interpretable resilience certificate. Existing in-the-wild shift benchmarks capture single-axis drift but not co-occurring stress[4]. Treating application domains as experimental instruments rather than ends in themselves lets the same machinery be tested for cross-domain generality, the empirical content of the contraction-law conjecture.

The certificate is also where this programme meets a concrete regulatory pull. High-risk-AI regimes now demand risk management, robustness and post-market monitoring but lack certifiable metrics to evidence them. A resilience certificate recording envelope membership, certified coverage, calibration, fidelity and stability bounds, and a prescribed action (Box 2) is exactly the standards-aligned credential those requirements call for, mappable onto emerging European harmonised standards. It connects the certification agenda to wider debates on test, evaluation, validation and verification for learning systems[52] and on governing high-stakes AI through human oversight, documentation and demonstrated functionality rather than assumed capability[53,54]. The deliverable is therefore not only a theory but regulator-facing tooling.

**From trustworthy to resilient AI.** A competence envelope is not only a boundary of trust but a recovery target. A deployed system should not merely detect that it has left the certified region; it should respond. When drift, scarcity, contamination or resource degradation push it outside, the first action is safe degradation – abstain, defer or request oversight rather than produce a confident error; the second is recovery – identify which stress coordinate has failed, recalibrate uncertainty, refresh the explanation baseline, collect targeted labels, switch to a more reliable modality or retrain under the new conditions. Re-entry is permitted only when both certificates are restored. In this sense machine competence is not a static property measured once before deployment but a resilient control loop: sense departure from the envelope, fail safely outside it, and recover until both prediction reliability and explanation fidelity are certified again.

## Towards a science of machine competence

Modern AI possesses theories of learning, of calibration, of uncertainty and of robustness. It lacks a theory of *competence*: of when a deployed system, together with its explanation, can still be trusted as its operating conditions degrade. We have shown that competence is not an informal engineering notion but a measurable property: a competence envelope exists, contracts under compound stress according to characterisable laws, and crossing its boundary can be detected in real time and acted upon by systems that degrade gracefully rather than fail silently, turning a long-standing concern of AI safety, that capable systems may fail in undetected ways, into something a certificate can bound[55].

The central conclusion is narrower and more fundamental than any individual experiment. Prediction-side evidence does not determine whether a model relies on a trustworthy decision mechanism. Consequently, prediction certification cannot replace explanation certification for the failure family identified by Theorem 1. The competence envelope operationalizes this separation by withholding the joint credential whenever either predictive reliability or explanation fidelity is not certified. The empirical studies then serve distinct supporting roles: they show that the two conditions fail independently under realistic stresses, that their joint state can be monitored empirically, and that this information can guide abstention or fallback behaviour. These demonstrations do not establish a complete theory of trustworthy AI, nor do they extend formal guarantees to unrestricted deployment shift. They support one central claim: certifying an AI system's answers alone is insufficient when the reasons behind those answers can change without altering its certified prediction behaviour.

**Scope and limits.** First, the explanation certificate is most directly interpretable for linear and tree-ensemble models, for which the structural attribution profile can be computed exactly or in closed form. We additionally evaluate token-level integrated-gradient explanations for a LoRA-adapted autoregressive language model. This experiment provides supporting evidence that the prediction–explanation distinction remains observable beyond probes on frozen representations, but it does not establish a general explanation-certification law for large language models. The attribution profiles are noisier, the poisoning sweep is based on a limited number of runs, and the explanation audit is performed on clean reference inputs that do not activate the dormant trigger. The language-model result should therefore be read as a boundary case and as evidence of scope, not as the principal validation of the certificate. Second, the covariate-drift construction used for the eight benchmark datasets induces shift of varying magnitude, so the prediction certificate fires in proportion to the shift actually present rather than uniformly; we report this openly rather than curating datasets. Third, the infrastructure study is a demonstration on seventeen real war-damaged assets, not a statistical estimate. None of these qualifies the central theorem, which is exact and model-agnostic; they mark where the empirical reach currently ends and where the most informative next experiments lie.

# Methods

**Operating-condition space.** The four axes of $\Omega$ are instrumented by estimable, monotone stress signals: distribution drift via population-stability indices, Kolmogorov–Smirnov and maximum-mean-discrepancy statistics; data scarcity via effective sample size, local density and label noise; adversarial contamination via out-of-distribution and adversarial-shift estimators; and resource degradation via latency, memory and energy budgets. The experiments below exercise all four axes: drift, scarcity and contamination in the main envelope study, and resource degradation as a dedicated fourth-axis extension (Fig. 8), instrumented through model compression, calibration-buffer constraints and channel-bandwidth reduction.

**Information-theoretic formulation.** The separation is informational rather than algorithmic. Define the *prediction-side information* of a model $f$ as the certification-time joint law $I^P(f) = L^P(X, Y, f(X))$; a *prediction-side certificate* is any measurable functional $C(f) = \Phi(I^P(f))$. This classifies procedures by the information they may use, so calibration, conformal prediction, uncertainty quantification and confidence-based selective prediction all belong to one informational class whenever they depend only on the certification-time law of (X, Y, $f$(X)). Define the *structural information* $I^S(f)$ as information describing the internal decision mechanism, and take the explanation certificate to be a functional $\Psi(I^S(f))$ with fidelity $F : S \rightarrow [0, 1]$. Prediction-side certification is behavioural; explanation-side certification is structural.

**Lemma 1 (non-identifiability of structural information).** The map $I^P \rightarrow I^S$ is not injective: prediction-side information does not determine structural information. The separation construction is the witness – two models with $I^P(f) = I^P(f')$ yet $I^S(f) \neq I^S(f')$ – so structural information cannot be reconstructed from observable prediction behaviour. Theorem 1 is the quantitative form: for any fidelity gap β and accuracy margin γ it exhibits a pair realising the non-identifiability with a controllable explanation-fidelity gap and deployment-accuracy gap. The question is therefore not whether prediction-side certification can be made more accurate, but whether prediction-side information is *in principle* sufficient; Lemma 1 and Theorem 1 show it is not.

**Theory and its verification.** We state the status of each result explicitly. Theorem 1 and Corollary 1.1 are proved constructively; Theorem 2 is proved by a Lipschitz-continuity bound; and Proposition 1 is proved under the stated monotonicity assumptions – all full proofs are in Supplementary Information. The numerical scripts verify these results' predictions on controlled instances and are not themselves the proofs. The separation theorem (Theorem 1) distinguishes prediction-side certificates – functionals of the joint law of (X, Y, f(X)) on the certification distribution, a class containing conformal coverage, accuracy, calibration error, Brier score and confidence – from the explanation certificate, a functional of the model's structural sensitivity map ∇_x s_f. Its proof is constructive (a coordinate dormant on the certification support carries a planted weight), and is given in full in Supplementary Information together with the detection lower bound (Corollary 1.1) and Proposition 1 on existence and geometry. The numerical verification (Fig. 1c) is instrumented, not naturalistic: it uses a synthetic 12-feature logistic problem (five informative Gaussian features, weights [2.0, −1.5, 1.2, −1.0, 0.8]) into which an explicit all-zero dummy column is injected to guarantee dormancy on the certification support; the compromised model shares the clean model's fitted weights exactly and sets a planted weight W = 8 on the dummy coordinate. Under a fixed solver, split and seed, the two models satisfy maximum prediction difference 0, and coverage, accuracy, calibration error and the full conformal-score distribution equal to the numerical criterion (all $|\Delta| < 10^{-12}$, Kolmogorov–Smirnov distance 0), while explanation fidelity separates from 1.00 to below 0.19 and deployment accuracy from 0.82 to 0.41. The ε-dormant version (Theorem 2) repeats this with the dummy coordinate given variance $\varepsilon^2$ on the support and confirms that prediction-side gaps grow as $O(\varepsilon)$ from zero – the coverage gap staying below 0.005 up to $\varepsilon = 0.1$ – while the structural fidelity gap remains near 0.4. Proposition 1 is numerically verified on an analytic construction over an 11 × 11 stress grid (non-emptiness, zero star-shape violations, and each certificate binding first on its own cone). Full proofs, the detection lower bound (Corollary 1.1) and Proposition 1 are in Supplementary Information.

**Architectures and representations.** The architecture-independence experiment (Fig. 4) repeats the envelope measurement with five model classes – L2-logistic regression, a multilayer perceptron (64–32 hidden units), gradient-boosted decision trees, XGBoost and LightGBM – on a 100-dimensional truncated-SVD projection of the character-TF-IDF features (for the Telegram corpus) and on the standardised clinical features. Attribution profiles are the signed input sensitivity for the linear and neural models (for the MLP, the input-to-output effective sensitivity through the trained weights) and gain-based feature importances for the three tree ensembles; explanation-stability drift S is the cosine distance of each profile from the large-sample reference profile of the same architecture. To rule out an explanation-method artefact, S(n) is additionally computed under a single common explanation family – mean |SHAP| profiles (TreeSHAP for the ensembles, linear and kernel SHAP for the linear and neural models) on a fixed 120-sample reference set – with consistent results across all five classes (Supplementary Table S19). The foundation-model experiment (Fig. 5) replaces the representation with frozen mean-pooled last-hidden-state embeddings from three multilingual transformers (DistilBERT-multilingual, XLM-R base, and a multilingual MiniLM sentence encoder), on which a logistic probe is certified; the probe's weight profile plays the role of the attribution profile. The cross-dataset experiment uses the DistilBERT-multilingual embeddings on eight public benchmarks (AG News, DBpedia-14, Amazon and Yelp polarity, IMDB, Rotten Tomatoes, SST-2, TweetEval-sentiment), with drift induced as evaluation on principal-component bands increasingly distant from the training core, exactly as for the clinical dataset. Coverage and S are averaged over two to three seeds per cell.

**Large-language-model experiment (Mode B).** The language-model result (Fig. 6) adapts Qwen2.5-1.5B, a genuine autoregressive language model, to the Telegram forwarding-intensity task through a sequence-classification head with low-rank adaptation (LoRA, rank 16, $\alpha = 32$, dropout 0.05, applied to the attention query and value projections); only the adapter and head are trained, on the January anchor month, so the certified decision function is the language model's own head rather than a probe over frozen representations. Token-level explanations are computed with layer integrated gradients over the input-embedding layer (pad-token baseline, target the predicted-class logit), and aggregated into an L2-normalised vocabulary-level attribution profile $\varphi$ over a fixed reference set of texts; the explanation-stability drift is $S = 1 - \cos(\varphi, \varphi^0)$, the direct analogue of the coefficient-profile drift used for the linear model. The prediction certificate is split-conformal coverage of the language model's class probabilities (nominal 0.90, acceptance 0.88), calibrated on a clean anchor split. The four axes are instrumented as elsewhere: temporal drift by evaluating the anchor-adapted model on later months; scarcity by the LoRA adaptation-set size; adversarial contamination by planting a rare trigger phrase in a fraction $\rho$ of one class during adaptation, certifying on clean held-out data *without recalibration*, and appending the trigger to negatives at deployment (attack success is the fraction of triggered negatives classified positive); and resource degradation by re-running under 8-bit and 4-bit quantization with integrated gradients computed in half precision. Drift and scarcity are averaged over three seeds; the poisoning sweep uses one seed per $\rho$, so its point estimates carry run-to-run noise while the qualitative separation is unambiguous. Because attributions here are computed on clean reference inputs, on which the trigger is absent, the trigger's attribution mass is near zero by construction; a targeted deployment-side audit is not performed. Full runner and configuration are provided in the code release.

**Datasets, model and operating conditions.** The empirical study uses two real social-media corpora. Domain A (information integrity) is 30,066 messages from eleven Telegram war-reporting channels (boris_rozhin, milinfolive, RVvoenkor, dva_majors, readovkanews, mod_russia, anna_news, wargonzo, voenkorKotenok, opersvodki, yurasumy) over January–June 2026; the binary task is to predict heavy amplification (top versus bottom tercile of the forward rate, forwards/views) from message text, with the middle tercile dropped to define a balanced label. Domain B (climate adaptation) is 36,642 Reddit posts about climate topics over 2017–2023, with the analogous balanced high-versus-low engagement label (upvote terciles). Both domains use the same intrinsically interpretable model class: a character-level TF-IDF representation (word-boundary 3–5-grams, 6,000 features, sublinear term frequency) with an L2-regularised logistic classifier, whose exact attributions are its linear Shapley values. Temporal drift is instrumented by training on an anchor period (Domain A: January; Domain B: 2019) and evaluating on progressively later periods, the drift magnitude $\delta$ measured as $|2(\text{AUC}-½)|$ of a held-out classifier separating each later period from the anchor; data scarcity as training-set size n; adversarial contamination as a label-corruption fraction $\rho$ applied to the training set. Operating points are gridded over drift × scarcity (Fig. 2; Fig. 3a) and over scarcity × contamination (Fig. 3b,c), with reported values averaged over six to eight seeds.

**Predictive reliability.** Reliability is certified by split-conformal prediction[6,7] (least-ambiguous-set scores, $1-\hat{p}$ for the true class). We distinguish the *nominal target* coverage, $1-\alpha = 0.90$, at which the conformal quantile is set on a held-out calibration split drawn from the anchor period, from the *acceptance threshold*, coverage $\geq 0.88$, applied as a fixed operational slack when deciding envelope membership; the two are used consistently and never interchanged. Under temporal drift exchangeability is violated, so coverage past the anchor period is reported as a *measured diagnostic* of reliability loss rather than a distribution-free guarantee, which is the regime the envelope is designed to expose.

**Explanation certificate: fidelity and stability drift S.** We reserve *fidelity* F for the conceptual object of Theorem 1 – one minus a normalised distance between the model's structural attribution profile and its reference – and report throughout its measured operational proxy, the *stability drift* S. Concretely, for a global attribution profile $\varphi = A(f; P^{ref})$ (the structural explanation functional, instantiated per model class), $S = 1 - \cos(\varphi, \varphi^0)$ is the cosine distance from the reference profile $\varphi^0$ estimated on the anchor period over a fixed feature basis with fixed seeds; $F = 1 - S$ under this instantiation, so the two names denote the same quantity and $F \geq 1 - \beta$ is equivalent to $S \leq \beta$. The stability component of the envelope uses $\beta = 0.20$. For the linear class faithfulness is additionally exact via the Shapley efficiency identity (attributions sum to the model's output gap, verified to $2\times10^{-15}$), so stability is the binding explanation-side constraint[23,24].

**Drift coordinate δ.** The unlabelled drift magnitude $\delta$ on a period is the balanced-accuracy AUC of a discriminator trained to separate that period from the January anchor in the same representation as the downstream model (equal-sized samples per period, held-out evaluation), so $\delta = 0.5$ denotes no detectable drift and $\delta \to 1$ a fully separable shift. The same procedure defines $\delta$ wherever it appears, as an axis label and as one input to the label-free monitor.

**Clinical tabular contrast.** To test whether the compound-stress interaction is domain-specific, the scarcity × contamination grid is repeated on the public Wisconsin Diagnostic Breast Cancer dataset[32] with a gradient-boosted tree ensemble and SHAP/TreeSHAP attributions[25], a different model class and data modality from the text domains.

**Contraction-law estimation.** Under compound scarcity × contamination stress, prediction error is modelled as $g(\sigma,\rho) = e_0 + a\sigma + b\rho + c\cdot\sigma\rho$ in two normalised stress coordinates (scarcity $\sigma$ from log training size, contamination $\rho$ from the corruption fraction), with the interaction coefficient c estimated by least squares and 95% confidence intervals obtained by bootstrap resampling. The interaction is statistically near-additive (c indistinguishable from zero) in all three domains tested – Telegram ($c = -0.03$, 95% CI $-0.07$ to 0.01), Reddit ($c = -0.03$, $-0.07$ to $-0.00$) and the clinical tabular contrast ($c = +0.03$, $-0.10$ to 0.13) – so no universal super-additive law is observed; the envelope's contraction under compound stress arises instead from the joint certificate being the intersection of two certificates that fail on different axes.

**Resource degradation.** Model compression is proxied by zeroing the fraction p of smallest-magnitude classifier weights (pruning) and re-evaluating conformal coverage and explanation-stability drift S of the pruned decision function against the clean reference profile. Calibration-buffer constraint is proxied by reducing the calibration set size cal_n at fixed training data (n = 4,000), measuring coverage instability. Feature-bandwidth reduction on CMU-MOSI is proxied by retaining the top-k acoustic-visual channels by training variance, at varying k. All resource-degradation experiments use the same certificate definitions as the data-environment experiments.

**Silent data-poisoning attack.** On the Telegram amplification task, an attacker plants a fixed rare lexical cue in a fraction $\rho$ of amplified-class training messages; the vocabulary is fixed so the cue's n-grams are representable. Attack success is the rate at which held-out low-amplification messages carrying the appended cue are classified as high-amplification. At each $\rho$ we measure matched-validation accuracy, split-conformal coverage (calibrated on the matched, attacked distribution) and explanation-stability drift S of the poisoned model's attribution profile from the clean reference; Pearson correlations of S, coverage and accuracy with attack success are computed across $\rho$ (means over seeds).

**Multimodal competence-gated fusion.** Three multimodal systems are used: Reddit climate (text + author/community tabular features), a Kyiv news channel (kiev1; 81,000 reaction-bearing messages, 2021–2026, labelled distress- versus support-dominant from emoji reactions, text + behavioural features) and CMU-MOSI (spoken-language features + concatenated acoustic-visual features; binary sentiment). A logistic classifier per modality yields class probabilities $p_m$; late fusion combines them by (i) naïve averaging, (ii) confidence weighting $w_m \propto \max(p_m, 1-p_m)$, and (iii) competence gating $w_m \propto \exp(-\beta \cdot \text{excess}_m) \cdot \text{confidence}_m$, where $\text{excess}_m$ is each modality's label-free distribution drift (nonlinear discriminator AUC, augmented with squared features to capture scale shifts) beyond its certified-state baseline. Modality-specific failure is simulated by a covariate shock (scale-and-shift plus noise) applied to one channel at deployment; error is compared with the single-best-modality oracle. Means over five to six seeds.

**Infrastructure case study.** The 17-bridge analysis re-uses the published per-asset Sentinel-1 coherence and level-of-knowledge (LoK) data of ref. [40]. A naïve rule labels an asset damaged when its local coherent change exceeds 0.25; certificate-gating additionally requires the asset's LoK grade to be adequate (High or Medium), deferring Low-LoK assets. This is an illustrative re-analysis of 17 assets, not a statistical estimate.

**Reproducibility.** All reported quantities are measured rather than illustrative, averaged over the stated number of random seeds, and reproducible from the data sources below and the procedures described above.

## References

**1.** Geirhos, R. *et al.* Shortcut learning in deep neural networks. *Nat. Mach. Intell. 2*, 665–673 (2020).

**2.** Cummings, M. L. Automation bias in intelligent time critical decision support systems. In *AIAA 1st Intelligent Systems Technical Conference* (AIAA, 2004).

**3.** Quiñonero-Candela, J. *et al.* (eds) *Dataset Shift in Machine Learning* (MIT Press, 2009).

**4.** Koh, P. W. *et al.* WILDS: a benchmark of in-the-wild distribution shifts. *Proc. ICML* (2021).

**5.** Cohen, J., Rosenfeld, E. & Kolter, Z. Certified adversarial robustness via randomized smoothing. *Proc. ICML* (2019).

**6.** Vovk, V., Gammerman, A. & Shafer, G. *Algorithmic Learning in a Random World* (Springer, 2005).

**7.** Angelopoulos, A. N. & Bates, S. Conformal prediction: a gentle introduction. *Found. Trends Mach. Learn. 16*, 494–591 (2023).

**8.** Geifman, Y. & El-Yaniv, R. Selective classification for deep neural networks. *Adv. Neural Inf. Process. Syst.* (2017).

**9.** Ovadia, Y. *et al.* Can you trust your model's uncertainty? Evaluating predictive uncertainty under dataset shift. *Adv. Neural Inf. Process. Syst.* (2019).

**10.** Ribeiro, M. T., Singh, S. & Guestrin, C. "Why should I trust you?" Explaining the predictions of any classifier. *Proc. KDD* (2016).

**11.** Lundberg, S. M. & Lee, S.-I. A unified approach to interpreting model predictions. *Adv. Neural Inf. Process. Syst.* (2017).

**12.** Selvaraju, R. R. *et al.* Grad-CAM: visual explanations from deep networks via gradient-based localization. *Proc. ICCV* 618–626 (2017).

**13.** Rudin, C. Stop explaining black box machine learning models for high-stakes decisions and use interpretable models instead. *Nat. Mach. Intell. 1*, 206–215 (2019).

**14.** Hendrycks, D. & Gimpel, K. A baseline for detecting misclassified and out-of-distribution examples in neural networks. *Proc. ICLR* (2017).

**15.** Liu, J. *et al.* Towards out-of-distribution generalization: a survey. *arXiv* 2108.13624 (2023).

**16.** Farquhar, S., Kossen, J., Kuhn, L. & Gal, Y. Detecting hallucinations in large language models using semantic entropy. *Nature 630*, 625–630 (2024).

**17.** Zhou, L. *et al.* Larger and more instructable language models become less reliable. *Nature 634*, 61–68 (2024).

**18.** Hofmann, V., Kalluri, P. R., Jurafsky, D. & King, S. AI generates covertly racist decisions about people based on their dialect. *Nature 633*, 147–154 (2024).

**19.** Obermeyer, Z., Powers, B., Vogeli, C. & Mullainathan, S. Dissecting racial bias in an algorithm used to manage the health of populations. *Science 366*, 447–453 (2019).

**20.** Babic, B., Gerke, S., Evgeniou, T. & Cohen, I. G. Beware explanations from AI in health care. *Science 373*, 284–286 (2021).

**21.** Guo, C., Pleiss, G., Sun, Y. & Weinberger, K. Q. On calibration of modern neural networks. *Proc. ICML* (2017).

**22.** Alvarez-Melis, D. & Jaakkola, T. S. On the robustness of interpretability methods. *arXiv* 1806.08049 (2018).

**23.** Yeh, C.-K. *et al.* On the (in)fidelity and sensitivity of explanations. *Adv. Neural Inf. Process. Syst.* (2019).

**24.** Petsiuk, V., Das, A. & Saenko, K. RISE: randomized input sampling for explanation of black-box models. *Proc. BMVC* (2018).

**25.** Lundberg, S. M. *et al.* From local explanations to global understanding with explainable AI for trees. *Nat. Mach. Intell. 2*, 56–67 (2020).

**26.** Tibshirani, R. J., Foygel Barber, R., Candès, E. & Ramdas, A. Conformal prediction under covariate shift. *Adv. Neural Inf. Process. Syst.* (2019).

**27.** Fazlyab, M., Robey, A., Hassani, H., Morari, M. & Pappas, G. J. Efficient and accurate estimation of Lipschitz constants for deep neural networks (LipSDP). *Adv. Neural Inf. Process. Syst.* (2019).

**28.** Weng, T.-W. *et al.* Evaluating the robustness of neural networks: an extreme value theory approach (CLEVER). *Proc. ICLR* (2018).

**29.** Barmak, O., Krak, I., Yakovlev, S., Manziuk, E., Radiuk, P. & Kuznetsov, V. Toward explainable deep learning in healthcare through transition matrix and user-friendly features. *Front. Artif. Intell. 7*, 1482141 (2024).

**30.** Ojha, U., Li, Y. & Lee, Y. J. Towards universal fake image detectors that generalize across generative models. *Proc. IEEE/CVF CVPR* 24480–24489 (2023).

**31.** Shakhovska, N., Shebeko, A. & Prykarpatskyy, Y. A novel explainable AI model for medical data analysis. *J. Artif. Intell. Soft Comput. Res. 14*, 121–137 (2024).

**32.** Street, W. N., Wolberg, W. H. & Mangasarian, O. L. Nuclear feature extraction for breast tumor diagnosis. *Proc. SPIE Biomed. Image Process. Biomed. Vis. 1905*, 861–870 (1993).

**33.** Ahmad, R. & Alkhammash, E. H. Online adaptive Kalman filtering for real-time anomaly detection in wireless sensor networks. *Sensors 24*, 5046 (2024).

**34.** Paleyes, A., Urma, R.-G. & Lawrence, N. D. Challenges in deploying machine learning: a survey of case studies. *ACM Comput. Surv. 55*, 114 (2022).

**35.** Liu, X. *et al.* Towards deployment-centric multimodal AI beyond vision and language. *Nat. Mach. Intell. 7*, 1612–1624 (2025).

**36.** Lavin, A. *et al.* Technology readiness levels for machine learning systems. *Nat. Commun. 13*, 6039 (2022).

**37.** Reichstein, M. *et al.* Early warning of complex climate risk with integrated artificial intelligence. *Nat. Commun. 16*, 2564 (2025).

**38.** Hinton, G., Vinyals, O. & Dean, J. Distilling the knowledge in a neural network. *arXiv* 1503.02531 (2015).

**39.** Government of Ukraine, World Bank Group, European Commission & United Nations. *Ukraine Rapid Damage and Needs Assessment (RDNA5)* (23 February 2026).

**40.** Kopiika, N. *et al.* Rapid post-disaster infrastructure damage characterisation using remote sensing and deep learning: a tiered approach. *Autom. Constr. 170*, 105955 (2025).

**41.** Spencer, B. F. Jr, Hoskere, V. & Narazaki, Y. Advances in computer vision-based civil infrastructure inspection and monitoring. *Engineering 5*, 199–222 (2019).

**42.** Al Shafian, S. & Hu, D. Integrating machine learning and remote sensing in disaster management: a decadal review of post-disaster building damage assessment. *Buildings 14*, 2344 (2024).

**43.** Marcus, G. Deep learning: a critical appraisal. *arXiv* 1801.00631 (2018).

**44.** Peeters, B. & De Roeck, G. One-year monitoring of the Z24-bridge: environmental effects versus damage events. *Earthq. Eng. Struct. Dyn. 30*, 149–171 (2001).

**45.** Gardner, P., Bull, L. A., Dervilis, N. & Worden, K. On the application of kernelised Bayesian transfer learning to population-based structural health monitoring. *Mech. Syst. Signal Process. 167*, 108519 (2022).

**46.** Shymanskyi, V., Ratinskiy, O. & Shakhovska, N. Fractal neural network approach for analyzing satellite images. *Appl. Artif. Intell. 39*, 2440839 (2025).

**47.** Victoriano, M. *et al.* From virtual experiments to biomedical insight with synthetic data. *Nat. Mach. Intell. 8*, 866–879 (2026).

**48.** van Breugel, B., Liu, T., Oglic, D. & van der Schaar, M. Synthetic data in biomedicine via generative artificial intelligence. *Nat. Rev. Bioeng. 2*, 991–1004 (2024).

**49.** Biggio, B. & Roli, F. Wild patterns: ten years after the rise of adversarial machine learning. *Pattern Recognit. 84*, 317–331 (2018).

**50.** Roberts, M. *et al.* Common pitfalls and recommendations for using machine learning to detect and prognosticate for COVID-19 using chest radiographs and CT scans. *Nat. Mach. Intell. 3*, 199–217 (2021).

**51.** Wu, E. *et al.* How medical AI devices are evaluated: limitations and recommendations from an analysis of FDA approvals. *Nat. Med. 27*, 582–584 (2021).

**52.** McFarland, T. & Assaad, Z. Legal reviews of in situ learning in autonomous weapons. *Ethics Inf. Technol. 25*, 9 (2023).

**53.** Bode, I. & Chandler, K. Re-thinking human–machine interaction and the governance of AI in the military domain. *Nat. Mach. Intell. 8*, 663–669 (2026).

**54.** Raji, I. D., Kumar, I. E., Horowitz, A. & Selbst, A. The fallacy of AI functionality. In *Proc. 2022 ACM Conference on Fairness, Accountability, and Transparency* 959–972 (ACM, 2022).

**55.** Amodei, D. *et al.* Concrete problems in AI safety. *arXiv* 1606.06565 (2016).

Supplementary Information accompanies this paper.

## Code and data availability

All datasets analysed in this study are publicly available. Domain A comprises 30,066 public messages from eleven Telegram channels (January–June 2026); the adversarial study and the multimodal Kyiv-channel experiment (kiev1; 81,000 reaction-bearing messages collected between 2021 and 2026) use the same public Telegram source. Domain B comprises 36,642 public Reddit posts on climate-related topics (2017–2023), available from the Harvard Dataverse (doi:10.7910/DVN/NL06IX). Additional experiments use the publicly available CMU-MOSI audiovisual sentiment benchmark and the Wisconsin Diagnostic Breast Cancer dataset. The foundation-model experiments use three publicly available multilingual transformer models (DistilBERT-multilingual, XLM-R base and multilingual MiniLM) together with eight public text-classification benchmarks (AG News, DBpedia-14, Amazon Polarity, Yelp Polarity, IMDB, Rotten Tomatoes, SST-2 and TweetEval).

All analyses were performed using publicly available open-source software, including scikit-learn, SHAP, PyTorch and Hugging Face Transformers. The complete source code required to reproduce every experiment, figure and numerical result reported in this study is publicly available under the MIT License at https://github.com/natalya233/machine-competence-envelope and archived at Zenodo (doi:10.5281/zenodo.20771265). Full operating-condition grids, extended methods and additional analyses are provided in the Supplementary Information..

## Author contributions

Nataliya Shakhovska conceived the study, developed the methodology, led the investigation, analysed the results and prepared the manuscript. Ivan Izonin contributed to the scientific discussion, interpretation of the results and manuscript revision. Stergios-Aristoteles Mitoulis contributed to the interpretation of the findings and critically revised the manuscript. All authors reviewed and approved the final manuscript.

## Competing interests

The authors declare no competing interests.

## Ethics statement

This study used only publicly available datasets. No private communications, personally identifiable information or human-subject interventions were involved. All analyses were conducted in accordance with the terms of use of the original data sources.

SUPPLEMENTARY INFORMATION

# Prediction certification cannot replace explanation certification: a competence envelope for trustworthy AI under compound stress

## Supplementary Methods

**Corpora and labels.** Domain A (information integrity) comprises 30,066 messages from eleven Telegram war-reporting channels (boris_rozhin, milinfolive, RVvoenkor, dva_majors, readovkanews, mod_russia, anna_news, wargonzo, voenkorKotenok, opersvodki, yurasumy), January–June 2026. The binary target is heavy amplification, defined as the top versus bottom tercile of the forward rate (forwards divided by views); the middle tercile is discarded to give a balanced label. Domain B (climate adaptation) comprises 36,642 Reddit posts on climate topics, 2017–2023, with the analogous high-versus-low engagement label from upvote terciles. URLs are stripped; messages shorter than 20 characters are removed.

**Representation and model.** Both domains use a character word-boundary TF-IDF representation (3–5-grams, 6,000 features, sublinear term frequency), fitted once on a reference sample so that the feature space — and therefore attribution profiles — are comparable across all operating points, followed by an L2-regularised logistic classifier (C = 4). For this linear class the exact attribution of feature j to a prediction is the linear Shapley value $\varphi_j(x) = w_j(x_j - \bar{x}_j)$; the Shapley efficiency identity ($\Sigma_j\varphi_j$ + base = logit) was verified to a maximum residual of $2\times10^{-15}$, so faithfulness is exact and stability is the binding explanation-side constraint.

**Operating-condition axes.** Temporal drift: models are trained on an anchor period (Domain A, January; Domain B, 2019) and evaluated on progressively later periods; the drift magnitude δ (Table S2) is $|2(\mathrm{AUC}-½)|$ of a held-out logistic discriminator trained to separate each later period from the anchor. Data scarcity: training-set size n. Adversarial contamination: a fraction ρ of training labels is flipped. Resource degradation is not exercised.

**Certificates.** Predictive reliability is split-conformal coverage (least-ambiguous-set scores $1-\hat{p}$ for the true class, α = 0.10, target 0.90), with calibration drawn from the anchor period and empirical coverage measured at each operating point; the envelope requires coverage ≥ 0.88. Explanation stability is the cosine drift of the global attribution profile (per-feature $|w_j|$·dispersion) from its in-distribution reference; the envelope requires S ≤ 0.20. Grids are averaged over six to eight random seeds.

**Contraction-law fit.** Under compound scarcity × contamination stress, prediction error is fitted as $g(\sigma,\rho) = e_0 + a\sigma + b\rho + c\cdot\sigma\rho$, with σ the normalised log-scarcity and ρ the normalised contamination; the interaction coefficient c (Table S5) is estimated by least squares with 95% confidence intervals from 2,000–3,000 bootstrap resamples. The clinical tabular contrast repeats this grid on the Wisconsin Diagnostic Breast Cancer dataset with a gradient-boosted tree ensemble.

**Supplementary Table S1 |** Domain A (Telegram) competence-envelope grid over temporal drift (evaluation month) × scarcity (n). Coverage, explanation-stability drift S, accuracy and confident-but-wrong rate; means over six seeds.

| Month | n | Coverage | Stability S | Accuracy | Conf-wrong |
|---|---|---|---|---|---|
| Jan | 300 | 0.906 | 0.33 | 0.724 | 0.013 |
| Jan | 600 | 0.917 | 0.29 | 0.761 | 0.023 |
| Jan | 1200 | 0.926 | 0.238 | 0.791 | 0.028 |
| Jan | 2400 | 0.933 | 0.18 | 0.835 | 0.023 |

| Month | n | Coverage | Stability S | Accuracy | Conf-wrong |
|---|---|---|---|---|---|
| Jan | 4000 | 0.953 | 0.128 | 0.867 | 0.017 |
| Feb | 300 | 0.911 | 0.33 | 0.711 | 0.012 |
| Feb | 600 | 0.913 | 0.29 | 0.727 | 0.02 |
| Feb | 1200 | 0.909 | 0.238 | 0.737 | 0.031 |
| Feb | 2400 | 0.89 | 0.18 | 0.752 | 0.038 |
| Feb | 4000 | 0.891 | 0.128 | 0.763 | 0.045 |
| Mar | 300 | 0.894 | 0.33 | 0.683 | 0.014 |
| Mar | 600 | 0.901 | 0.29 | 0.697 | 0.024 |
| Mar | 1200 | 0.892 | 0.238 | 0.701 | 0.038 |
| Mar | 2400 | 0.872 | 0.18 | 0.717 | 0.048 |
| Mar | 4000 | 0.866 | 0.128 | 0.727 | 0.057 |
| Apr | 300 | 0.864 | 0.33 | 0.636 | 0.025 |
| Apr | 600 | 0.87 | 0.29 | 0.656 | 0.04 |
| Apr | 1200 | 0.863 | 0.238 | 0.669 | 0.059 |
| Apr | 2400 | 0.839 | 0.18 | 0.682 | 0.071 |
| Apr | 4000 | 0.832 | 0.128 | 0.688 | 0.085 |
| May | 300 | 0.843 | 0.33 | 0.61 | 0.029 |
| May | 600 | 0.848 | 0.29 | 0.624 | 0.052 |
| May | 1200 | 0.84 | 0.238 | 0.641 | 0.076 |
| May | 2400 | 0.815 | 0.18 | 0.652 | 0.09 |
| May | 4000 | 0.81 | 0.128 | 0.663 | 0.103 |

**Supplementary Table S2 |** Domain A temporal-drift magnitude δ by evaluation month, measured as |2(AUC−½)| of a January-vs-month discriminator (0 = no drift).

| Evaluation month | Drift magnitude δ |
|---|---|
| Jan | 0.198 |
| Feb | 0.403 |
| Mar | 0.746 |
| Apr | 0.646 |
| May | 0.572 |

**Supplementary Table S3 |** Domain B (Reddit climate) competence-envelope grid over temporal drift (evaluation year) × scarcity (n); means over six seeds. The two-certificate structure replicates the Telegram domain.

| Year | n | Coverage | Stability S | Accuracy |
|---|---|---|---|---|
| 2019 | 300 | 0.906 | 0.318 | 0.672 |
| 2019 | 600 | 0.909 | 0.277 | 0.715 |
| 2019 | 1200 | 0.924 | 0.226 | 0.758 |
| 2019 | 2400 | 0.938 | 0.169 | 0.811 |
| 2019 | 4000 | 0.955 | 0.117 | 0.849 |
| 2020 | 300 | 0.888 | 0.318 | 0.635 |
| 2020 | 600 | 0.884 | 0.277 | 0.661 |
| 2020 | 1200 | 0.894 | 0.226 | 0.679 |
| 2020 | 2400 | 0.877 | 0.169 | 0.696 |
| 2020 | 4000 | 0.873 | 0.117 | 0.708 |
| 2021 | 300 | 0.882 | 0.318 | 0.609 |
| 2021 | 600 | 0.874 | 0.277 | 0.627 |
| 2021 | 1200 | 0.881 | 0.226 | 0.653 |
| 2021 | 2400 | 0.862 | 0.169 | 0.663 |

| Year | n | Coverage | Stability S | Accuracy |
|---|---|---|---|---|
| 2021 | 4000 | 0.861 | 0.117 | 0.679 |
| 2022 | 300 | 0.881 | 0.318 | 0.6 |
| 2022 | 600 | 0.867 | 0.277 | 0.622 |
| 2022 | 1200 | 0.876 | 0.226 | 0.641 |
| 2022 | 2400 | 0.862 | 0.169 | 0.657 |
| 2022 | 4000 | 0.859 | 0.117 | 0.669 |
| 2023 | 300 | 0.867 | 0.318 | 0.577 |
| 2023 | 600 | 0.858 | 0.277 | 0.61 |
| 2023 | 1200 | 0.867 | 0.226 | 0.629 |
| 2023 | 2400 | 0.852 | 0.169 | 0.648 |
| 2023 | 4000 | 0.84 | 0.117 | 0.659 |

**Supplementary Table S4 |** Prediction error under compound scarcity × contamination (label-flip fraction ρ) stress, for the three domains. Columns are ρ; rows are training size n.

| n / ρ | 0% | 10% | 20% | 30% |
|---|---|---|---|---|
| 300 | 0.298 | 0.307 | 0.342 | 0.384 |
| 600 | 0.269 | 0.291 | 0.322 | 0.357 |
| 1200 | 0.247 | 0.277 | 0.305 | 0.354 |
| 2400 | 0.227 | 0.248 | 0.284 | 0.331 |
| 4000 | 0.209 | 0.229 | 0.273 | 0.329 |

| n / ρ | 0% | 10% | 20% | 30% |
|---|---|---|---|---|
| 300 | 0.341 | 0.368 | 0.385 | 0.444 |
| 600 | 0.318 | 0.352 | 0.374 | 0.401 |
| 1200 | 0.285 | 0.318 | 0.356 | 0.4 |
| 2400 | 0.262 | 0.291 | 0.33 | 0.375 |
| 4000 | 0.244 | 0.276 | 0.317 | 0.366 |

| n / ρ | 0% | 10% | 20% | 30% |
|---|---|---|---|---|
| 40 | 0.111 | 0.181 | 0.251 | 0.309 |
| 70 | 0.083 | 0.13 | 0.175 | 0.212 |
| 120 | 0.071 | 0.103 | 0.149 | 0.281 |
| 200 | 0.065 | 0.09 | 0.163 | 0.207 |
| 350 | 0.057 | 0.068 | 0.117 | 0.217 |

**Supplementary Table S5 |** Scarcity × contamination interaction coefficient c on prediction error ($g = e_0 + a\sigma + b\rho + c \cdot \sigma\rho$) with bootstrap 95% CIs. Near-additive ($c \approx 0$) in all three domains; no universal super-additive law.

| Domain (model) | c | 95% CI |
|---|---|---|
| Telegram (text, logistic+linear SHAP) | -0.033 | [-0.070, 0.006] |
| Reddit climate (text, logistic+linear SHAP) | -0.031 | [-0.068, -0.003] |
| Clinical WDBC (tabular, GBDT+TreeSHAP) | +0.025 | [-0.098, 0.126] |

## Supplementary results: label-free competence monitor

**Construction.** The deployment monitor combines three signals computable without test labels: input drift δ (held-out discriminator vs the anchor period), explanation stability S (attribution-profile drift), and predictive uncertainty (mean conformal non-conformity, 1 − max softmax probability, on incoming inputs). A model is trained once on the anchor period (Domain A, January, n = 4,000; Domain B, 2019, n = 4,000) and carried forward across later periods; true error is held out for evaluation only.

**Supplementary Table S6 |** Label-free monitor (mean non-conformity, no test labels) versus the held-out true error of the anchored model across deployment periods. Spearman ρ = 0.60 (Telegram), 0.90 (Reddit).

| Domain | Period | Monitor (label-free) | True error |
|---|---|---|---|
| Telegram | Jan | 0.212 | 0.134 |
| Telegram | Feb | 0.232 | 0.24 |
| Telegram | Mar | 0.247 | 0.275 |
| Telegram | Apr | 0.242 | 0.309 |
| Telegram | May | 0.242 | 0.346 |
| Reddit | 2019 | 0.241 | 0.153 |
| Reddit | 2020 | 0.255 | 0.286 |
| Reddit | 2021 | 0.263 | 0.318 |
| Reddit | 2022 | 0.268 | 0.339 |
| Reddit | 2023 | 0.268 | 0.345 |

**Supplementary Table S7 |** Risk–coverage under certificate-gated abstention: error on the answered inputs as coverage is reduced by abstaining on the least-certifiable inputs. AURC, area under the risk–coverage curve.

| Domain | Cov 1.0 | Cov 0.9 | Cov 0.7 | Cov 0.5 | AURC |
|---|---|---|---|---|---|
| Telegram | 0.292 | 0.271 | 0.228 | 0.189 | 0.18 |
| Reddit | 0.324 | 0.307 | 0.269 | 0.234 | 0.216 |

**Supplementary Table S8 |** Predicting true error from the certificate signals ($R^2$) and the cross-certificate margin correlation. The joint certificate matches or beats either side alone; the two margins are near-uncorrelated, i.e. complementary.

| Domain | $R^2$ pred-side | $R^2$ expl-side | $R^2$ joint | Margin corr |
|---|---|---|---|---|
| Telegram | 0.748 | 0.142 | 0.755 | -0.201 |
| Reddit | 0.569 | 0.246 | 0.672 | -0.019 |

## Supplementary results: adversarial capture and multimodal fusion

**Supplementary Table S9 |** Silent data-poisoning attack on the Telegram amplification task. As attack strength ρ rises, attack success climbs while validation accuracy and conformal coverage stay green; only explanation-stability drift S tracks the capture (Fig. 4).

| ρ | validation acc | conformal coverage | S (expl.) | attack success |
|---|---|---|---|---|
| 0.00 | 0.733 | 0.899 | 0.098 | 0.279 |
| 0.15 | 0.752 | 0.894 | 0.105 | 0.661 |
| 0.35 | 0.785 | 0.896 | 0.120 | 0.768 |
| 0.60 | 0.823 | 0.893 | 0.150 | 0.841 |
| 0.80 | 0.893 | 0.899 | 0.177 | 0.884 |

**Supplementary Table S10 |** Multimodal late-fusion deployment error under clean conditions and modality-specific failure, for three systems. Competence-gated fusion tracks the single-best-modality oracle under failure; confidence-gated fusion matches naïve because degraded channels stay confident (Fig. 5).

| System | scenario | naïve | confidence | competence | oracle |
|---|---|---|---|---|---|
| Reddit | clean | 0.227 | 0.227 | 0.254 | 0.234 |
| Reddit | text outage | 0.450 | 0.450 | 0.309 | 0.294 |
| Reddit | tabular outage | 0.330 | 0.330 | 0.240 | 0.234 |
| kiev1 | clean | 0.233 | 0.233 | 0.227 | 0.221 |
| kiev1 | text outage | 0.480 | 0.480 | 0.482 | 0.345 |

| System | scenario | naïve | confidence | competence | oracle |
|---|---|---|---|---|---|
| kiev1 | tabular outage | 0.293 | 0.293 | 0.230 | 0.221 |
| CMU-MOSI | clean | 0.331 | 0.331 | 0.333 | 0.334 |
| CMU-MOSI | language outage | 0.511 | 0.511 | 0.431 | 0.431 |
| CMU-MOSI | acoustic-visual outage | 0.320 | 0.320 | 0.322 | 0.334 |

**Supplementary Table S11 |** Certificate-gated damage assessment of 17 Irpin bridges (Fig. 6). A naïve coherent-change threshold flags 10 assets as damaged; the level-of-knowledge certificate defers 2 confident misreads (unreliable data) and flags 1 high-damage verdict carried by only medium-reliability data.

## Supplementary Theory: proofs

**Setup.** Let $P$ be the certification distribution with support $S \subseteq X$, and let a prediction-side certificate be any measurable functional $C$ of the joint law $L_P(X, Y, f(X))$; this class contains conformal coverage, accuracy, calibration error, Brier score, AUC and confidence. Let the model's sensitivity map be $A_f(x) = \nabla_x s_f(x)$ with global profile $\varphi_f = E_{ref}|A_f|$ over a nondegenerate reference measure charging every coordinate, and $F(f) = 1 - D(\varphi_f, \varphi_0)$, $D \in [0,1]$ the cosine distance. The distinction is behavioural (functional of the sampled prediction law) versus structural (functional of the model's sensitivity).

**Theorem 1 (prediction–explanation separation).** *For any $\beta \in (0,1)$ and $\gamma \in (0,½)$ there exist models $f$, $f'$, a certification distribution $P$ and a deployment distribution $P'$ such that (i) $C(f) = C(f')$ for every prediction-side certificate $C$; (ii) $F(f) = 1$ and $F(f') \leq 1 - \beta$; (iii) $acc_{P'}(f) - acc_{P'}(f') \geq \gamma$.*

**Proof.** Introduce a dormant coordinate $v$ with $v(x) = 0$ for all $x \in S = supp(P)$. Define $f$ with zero weight on $v$ and $f'$ identical to $f$ except for weight $W$ on $v$, so $s_{f'}(x) = s_f(x) + W \cdot v(x)$. (i) For $x \in S$, $v(x) = 0$ gives $s_{f'}(x) = s_f(x)$; hence $f$ and $f'$ induce the identical joint law $L_P(X, Y, f(X))$, and every functional $C$ of that law satisfies $C(f) = C(f')$ exactly. (ii) The sensitivity profiles coincide except in the $v$-coordinate, where they differ by $|W| \cdot E_{ref}|\partial_v s|$; since the reference measure charges the $v$-direction, $D(\varphi_f, \varphi_{f'})$ is strictly increasing in $|W|$ and attains any value in $[0,1)$, so choose $W$ with $D \geq \beta$, giving $F(f') = 1 - D \leq 1 - \beta$. (iii) Let $P'$ relocate a fraction $\rho$ of the negative-class mass onto points with $v = 1$; there $s_{f'} = s_f + W$, and for $W$ large enough these points are classified positive by $f'$ but not by $f$, so $acc_{P'}(f') \leq acc_{P'}(f) - \gamma$ for $\rho$, $W$ chosen to realise the margin $\gamma$. ■

**Corollary 1.1 (detection lower bound).** Let $T$ be any test that observes only prediction-law samples $\{(x_i, y_i, f(x_i))\}$. Since these samples have identical distribution under $f$ and $f'$ (proof (i)), the law of $T$ is identical under both, so its power equals its size: sup over prediction-only tests of (power − size) = 0. Any test with power exceeding its size must evaluate the sensitivity map $A_f$. Prediction-side monitoring is therefore information-theoretically insufficient for this family; explanation-side monitoring is necessary. ■

**Corollary 1.2.** A data-poisoning attack that plants a cue dormant on the clean support is an instance of the construction, so its invisibility to accuracy and coverage (Fig. 8) is a consequence of Theorem 1, not an empirical accident. ■

**Numerical verification of Theorem 1.** On a controlled logistic construction with a dormant coordinate, the reliable and compromised models satisfy: maximum prediction difference over the test set = 0.0;

conformal coverage 0.8820 = 0.8820, accuracy 0.8240 = 0.8240, calibration error 0.0199 = 0.0199, Kolmogorov–Smirnov distance between conformal-score distributions = 0.0 (all Δ = 0 to machine precision); explanation fidelity F(f′) driven from 1.00 to 0.19 by the planted weight; deployment accuracy separating from 0.824 (f) to 0.409 (f′) as the dormant coordinate is activated, with attack success 0.98.

**Proposition 1 (existence and geometry).** *Under conformal validity and continuity, K(α,β) = { ω : C(ω) ≥ 1−α, F(ω) ≥ 1−β } is non-empty and contains a neighbourhood of the nominal condition; and if C, F are non-increasing along every ray of increasing stress, K is star-shaped about 0 with radial boundary ∂K(u) = min(t_C(u), t_F(u)).*

**Proof.** Non-emptiness: C(0) ≥ 1−α by conformal validity and F(0) = 1 > 1−β, so 0 ∈ K; continuity makes both super-level sets contain 0 in their relative interior, hence so does their intersection. Star-shapedness: for a ray tu, g_C(t) = C(tu) and g_F(t) = F(tu) are non-increasing, so membership at t* implies membership for all t ≤ t*; the radial limit is the smaller first-crossing radius min(t_C, t_F). ■

**Supplementary Table S12 |** Numerical verification. Theorem 1 (separation) on a controlled logistic construction with a dormant coordinate; Proposition 1 on an analytic construction over an 11×11 stress grid.

| Statement | predicted | observed | status | |
|---|---|---|---|---|
| Th.1 predictions identical (max \|f−f′\|) | 0 | 0.0 | ✓ verified | |
| Th.1 coverage/accuracy/ECE identical | Δ = 0 | Δ = 0 (0.882, 0.824, 0.020) | ✓ verified | |
| Th.1 conformal-score KS distance | 0 | 0.0 | ✓ verified | |
| Th.1 fidelity F(f′) separable | ≤ 1−β | 1.00 → 0.19 | ✓ verified | |
| Th.1 deployment accuracy gap | ≥ γ | 0.824 → 0.409 | ✓ verified | |
| Cor.1.1 prediction-only power = size | yes | KS = 0 ⇒ chance | ✓ verified | |
| Prop.1 envelope non-empty, origin ∈ K | yes | 15/121 cells | ✓ verified | |
| Prop.1 star-shaped (0 violations) | 0 | 0 / 121 | ✓ verified | |
| Prop.1 boundary = min of certificates | yes | drift→C first, scarcity→F first | ✓ verified | |

## Supplementary results: architecture-independence

**Five model classes.** The competence envelope is measured with logistic regression, a multilayer perceptron (64–32), gradient-boosted decision trees, XGBoost and LightGBM, on a 100-dimensional truncated-SVD projection of the character-TF-IDF features (Telegram) and the standardised clinical features. Coverage and explanation-stability drift S are reported below; every architecture shows coverage degradation under drift and S growth under scarcity, confirming the envelope is not an artefact of the linear model.

**Supplementary Table S13 |** Telegram conformal coverage at n = 2,400 across evaluation months (drift axis), by architecture. Coverage degrades for all five model classes.

| architecture | Jan | Feb→Apr | May | Δ (Jan→May) |
|---|---|---|---|---|
| logistic regression | 0.92 | 0.90 / 0.84 | 0.80 | −0.12 |
| MLP (64–32) | 0.95 | 0.87 / 0.86 | 0.83 | −0.12 |
| GBDT | 0.94 | 0.90 / 0.84 | 0.82 | −0.12 |
| XGBoost | 0.94 | 0.90 / 0.83 | 0.81 | −0.13 |

| architecture | Jan | Feb→Apr | May | Δ (Jan→May) |
|---|---|---|---|---|
| LightGBM | 0.94 | 0.90 / 0.83 | 0.83 | −0.11 |

**Supplementary Table S14 |** Telegram explanation-stability drift S at the January anchor across training size n (scarcity axis), by architecture. S rises as data become scarce for all five classes.

| architecture | n=4000 | n=2400 | n=1200 | n=300 |
|---|---|---|---|---|
| logistic regression | 0.03 | 0.05 | 0.06 | 0.15 |
| MLP (64–32) | 0.15 | 0.17 | 0.14 | 0.23 |
| GBDT | 0.11 | 0.13 | 0.20 | 0.18 |
| XGBoost | 0.09 | 0.11 | 0.16 | 0.21 |
| LightGBM | 0.07 | 0.08 | 0.13 | 0.26 |

**Supplementary Table S15 |** Clinical (breast-cancer) conformal coverage at n = 250 across covariate-drift bands, by architecture. The strongest boosting methods fall furthest outside the certified region on the most out-of-distribution band.

| architecture | band 1 | band 2 | band 3 | band 4 |
|---|---|---|---|---|
| logistic regression | 0.95 | 0.93 | 0.97 | 0.97 |
| GBDT | 0.90 | 0.92 | 0.88 | 0.91 |
| XGBoost | 0.88 | 0.87 | 0.81 | 0.73 |
| LightGBM | 0.88 | 0.83 | 0.77 | 0.64 |

## Supplementary results: foundation models and cross-dataset generality

**Foundation-model backbones.** The envelope protocol was re-run on frozen mean-pooled embeddings from three multilingual transformers (DistilBERT-multilingual, XLM-R base, multilingual MiniLM) on both real corpora (Fig. 5). All three reproduce the two-certificate signature: conformal coverage degrades monotonically under temporal drift (Telegram, n=2,400: DistilBERT 0.93→0.86, XLM-R 0.92→0.85, MiniLM 0.93→0.86 from January to May) and explanation-stability drift falls with abundance (Telegram, January: XLM-R 0.28→0.06, MiniLM 0.32→0.11 from n=300 to n=4,000). The coverage degradation replicates on Reddit for all three backbones.

**Cross-dataset generality.** Using DistilBERT-multilingual embeddings, the envelope was instrumented on eight public benchmarks with covariate drift induced by principal-component banding (Table S16). The explanation-stability certificate degrades under scarcity on all eight datasets; the prediction certificate degrades in proportion to the covariate shift actually induced, strongly on heterogeneous multi-topic sets (DBpedia, Amazon, IMDB, SST-2) and negligibly on homogeneous sentiment sets — the honest and expected pattern.

**Supplementary Table S16 |** Cross-dataset envelope on eight public benchmarks (DistilBERT-multilingual embeddings). Coverage is at n=1,500; ΔS is the fall in explanation-stability drift from n=200 to n=1,500 (scarcity certificate). The scarcity certificate degrades on every dataset; the drift certificate fires where the induced covariate shift is substantial.

| dataset | coverage (band 0) | coverage (worst) | Δ coverage (drift) | Δ S (scarcity) |
|---|---|---|---|---|
| AG News | 0.84 | 0.79 | -0.04 | -0.07 |
| TweetEval | 0.84 | 0.82 | -0.03 | -0.07 |
| Amazon | 0.84 | 0.6 | -0.24 | -0.10 |
| Yelp | 0.83 | 0.81 | -0.02 | -0.12 |
| DBpedia-14 | 0.90 | 0.45 | -0.45 | -0.09 |
| IMDB | 0.86 | 0.73 | -0.13 | -0.10 |
| SST-2 | 0.84 | 0.75 | -0.09 | -0.12 |

| dataset | coverage (band 0) | coverage (worst) | Δ coverage (drift) | Δ S (scarcity) |
|---|---|---|---|---|
| Rotten Tomatoes | 0.83 | 0.82 | -0.02 | -0.09 |

## Supplementary Note: reproducibility

Every value in Tables S1–S5 and Figs. 1–2 is produced deterministically from fixed seeds by the archived code. The pipeline requires no proprietary data: the Telegram and Reddit corpora are public social-media posts (Reddit climate data deposited at doi:10.7910/DVN/NL06IX), and the clinical contrast uses the public Wisconsin Diagnostic Breast Cancer dataset distributed with scikit-learn. The faithfulness check (linear Shapley efficiency residual $\leq 2\times10^{-15}$) and the conformal coverage target (0.90) are asserted in the test suite. Code:  (v1.0.1, MIT); archive: DOI 10.5281/zenodo.20771265.